\documentclass[10pt,twocolumn,letterpaper]{article}

\usepackage[pagenumbers]{wacv}   

\usepackage{bm}
\usepackage{threeparttable}
\usepackage{pifont}
\newcommand{\cmark}{\ding{51}}   

\graphicspath{{picture/}}

\definecolor{wacvblue}{rgb}{0.21,0.49,0.74}
\definecolor{linkblue}{rgb}{0.21,0.49,0.74}
\usepackage[pagebackref,breaklinks,colorlinks,allcolors=wacvblue]{hyperref}

\def\wacvPaperID{322}
\def\confName{WACV}
\def\confYear{2027}

\newcommand{\mainboth}{24.22}     
\newcommand{\mainmean}{24.31}      
\newcommand{\mainstd}{2.22}        
\newcommand{\ninetime}{9.1}        
\newcommand{\ninegs}{9.1}          
\newcommand{\ninemean}{21.47}      

\newcommand{\demoboth}{25.50}      

\newcommand{\teachmean}{26.50}     

\newcommand{\ablscratchfull}{16.4} 

\newcommand{\ourgs}{23.9}          
\newcommand{\retimegs}{128}        
\newcommand{\ourfps}{689}          
\newcommand{\peakoff}{2.12}        
\newcommand{\peakopt}{0.83}        
\newcommand{\ourtime}{43}          

\newcommand{\dgsG}{15.9}   \newcommand{\dgstime}{71}      
\newcommand{\gflowG}{128}  \newcommand{\gflowtime}{36}    
\newcommand{\fourdgsG}{20.1} \newcommand{\fourdgstime}{58} 
\newcommand{\ninevram}{1.0}     
\newcommand{\dgsvram}{6.4}      
\newcommand{\gflowvram}{7.1}    
\newcommand{\fourdgsvram}{6.0}  
\newcommand{\retimevram}{9.3}   

\newcommand{\gflowmean}{20.06}     

\newcommand{\fourdgsmean}{21.20}   

\newcommand{\dgsmean}{18.73}       

\newcommand{\dgsgap}{5.6}          

\newcommand{\ablscratch}{16.4}     
\newcommand{\ablgfa}{20.48}        
\newcommand{\ninenomirror}{16.00}  
\newcommand{\ninenomirrorpeak}{17.41}
\newcommand{\ninemirror}{24.76}    

\newcommand{\rawmb}{61.1}          
\newcommand{\tfsfortymb}{2.8}      
\newcommand{\tfsfortymbps}{16.8}   
\newcommand{\tfstwentymb}{1.4}     
\newcommand{\tfstwentymbps}{8.4}   
\newcommand{\fpsforty}{1312}       
\newcommand{\fpstwenty}{1400}      
\newcommand{\questfps}{87}         
\newcommand{\questxfer}{0.44}      
\newcommand{\questmbps}{58}        
\newcommand{\questxferfull}{2.3}   

\newcommand{\scratchopencase}{17.49} \newcommand{\scratchopencaseN}{59}  

\newcommand{\dnamean}{25.20}      
\newcommand{\dnags}{42.4}        
\newcommand{\dnatime}{51}      
\newcommand{\dnadgsmean}{23.29}   
\newcommand{\dnafourmean}{24.45}  

\InputIfFileExists{macros_auto}{}{}

\newcommand{\coderelease}{\par\vspace{3pt}\noindent\raggedright{\small Code:\ \url{https://github.com/Jingong-Chen/Amortized-Anchor-Refinement}}\par}

\newcommand{\mpExp}{Sec.~\ref{sec::exp}}
\newcommand{\mpNine}{Table~\ref{tab::ninescene}}
\newcommand{\mpTfs}{Table~\ref{tab::tfs}}
\newcommand{\mpQual}{Figure~\ref{fig::qualitative}}

\title{Amortized Anchor Refinement for Deployable\\ Continuous-Time 4D Gaussian Reconstruction}

\author{
Jingong Chen$^{1}$ \quad Qingwen Zhang$^{2}$ \quad Sanghyeon Jun$^{1}$\\[2pt]
Chulwoo Pack$^{1}$ \quad Kyle Gao$^{3}$ \quad Kwanghee Won$^{1}$\\[4pt]
{\small $^{1}$Department of Electrical Engineering \& Computer Science, South Dakota State University, Brookings, SD, USA}\\
{\small $^{2}$Department of Robotics, Perception and Learning, KTH Royal Institute of Technology, Stockholm, Sweden}\\
{\small $^{3}$Department of Built Environment, Aalto University, Espoo, Finland}\\[2pt]
{\tt\small \{jingong.chen, Sanghyeon.Jun\}@jacks.sdstate.edu}\\
{\tt\small \{Chulwoo.Pack, Kwanghee.Won\}@sdstate.edu \quad qingwen@kth.se \quad kyle.gao@aalto.fi}
}

\begin{document}
\maketitle

\providecommand{\coderelease}{}
\begin{abstract}
Continuous-time 4D reconstruction remains impractical on standalone XR headsets.
Per-scene optimization demands deployment-infeasible compute, and lower budgets cause collapse rather than degrade gradually.
Feed-forward prediction is fast, but struggle to recover scene-specific detail.
We present \textbf{Amortized Anchor Refinement}, which uses a frozen backbone to predict an initial Gaussian representation and a short optimization to specialize it under a fixed compute budget, with a capacity floor preserving representational density. A training-free stage then applies a \textbf{persistent-homology constraint} to prune unstable Gaussians while preserving topologically persistent structures, and streams the resulting trajectories directly as scene flow. On the Stage-Capture benchmark, \textbf{Amortized Anchor Refinement} achieves $\mainmean{\pm}\mainstd$~dB, while our deployment experiments demonstrate reconstruction within the target budget on a single consumer GPU and playback on a standalone XR headset.\coderelease
\end{abstract}



\section{Introduction}
\label{sec::intro}
Reconstructing a dynamic scene so that a viewer may observe it from arbitrary viewpoints at arbitrary times is a long-standing goal of immersive media~\cite{broxton2020immersive}, enabling applications such as virtual reality, telepresence, and free-viewpoint replay.
Recent advances have brought this goal within reach. Neural radiance fields~\cite{mildenhall2020nerf} established photorealistic novel-view synthesis, while subsequent work improved scalability to unbounded scenes and deformable content~\cite{muller2022instantngp,barron2022mipnerf360,pumarola2021dnerf,park2021nerfies,park2021hypernerf,fridovichkeil2023kplanes}.
More recently, 3D Gaussian Splatting~\cite{kerbl3Dgaussians} replaced implicit fields with explicit primitives, enabling real-time rendering and motivating a broad family of continuous-time dynamic Gaussian representations.

\begin{figure}[t]
\centering
\includegraphics[width=\linewidth]{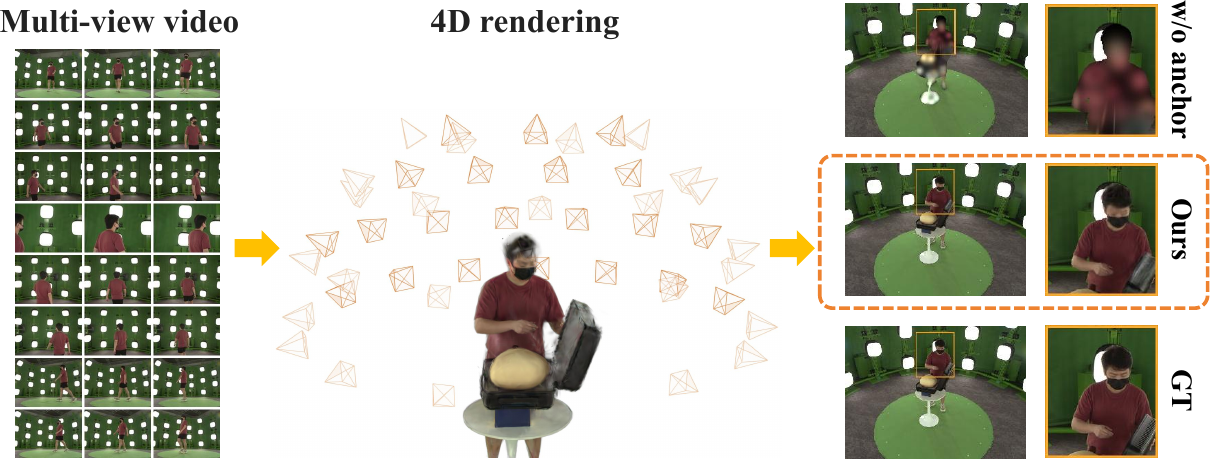}
\caption{\textbf{Amortized Anchor Refinement uses feedforward anchors as enabler.} Under the deployment budget of standalone XR headsets, optimizing a 4D Gaussian field from scratch collapses, whereas the same budget becomes sufficient for scene-specific refinement when initialized by a single feed-forward pass. Right: visual comparison of rendering at a held timestamp.}
\label{fig::motivation}
\end{figure}

Existing methods model dynamics either through deformation fields or explicit Gaussian trajectories~\cite{yang2024deformable,wu20244d,huang2024scgs,luiten2024dynamic,li2024spacetime,duan20244drotor,yang2024realtime4dgs,lee2024ex4dgs,retimegs}.
Given sufficient representational capacity and scene-specific optimization, these methods reconstruct complex real-world dynamics with high quality while answering queries at arbitrary times in real time.
However, this quality scales with the number of per-scene optimization iterations, whose cost places these methods out of reach of standalone XR headsets.

These headsets provide only a few gigabytes of memory, limited compute, and a wireless link for streaming.
Continuous-time reconstruction further tightens these constraints because the entire temporal field must remain resident to support arbitrary-time queries, making frame-by-frame streaming infeasible.
Existing compression methods reduce the size of an already-optimized field~\cite{fan2024lightgaussian,lu2024scaffold,lee2024compact3dgs}, while streaming systems transmit frame-to-frame residuals~\cite{threedgstream,queen,videogs}, but neither asks how the field itself should be learned when the deployment budget of standalone XR headsets is imposed from the outset.

A direct solution is to reduce the optimization budget by using fewer Gaussians and fewer iterations. However, this fails because photometric gradients provide no consistent update direction for unanchored Gaussians: the densify-and-prune schedule removes them faster than it can establish valid geometry, causing reconstruction quality to collapse to $\ablscratchfull$~dB rather than degrade gracefully to a usable solution. Thus, under a constrained budget, optimization is dominated by geometry discovery rather than scene refinement.

Feed-forward prediction avoids this search entirely. Recent amortized methods reconstruct static~\cite{dust3r,vggt,charatan2024pixelsplat,chen2024mvsplat,tang2024lgm} and dynamic scenes~\cite{movies,l4gm,xu20254dgt} within seconds, but their accuracy remains constrained by the learned geometric prior and does not fully capture scene-specific detail. These limitations are complementary: feed-forward prediction provides an initial geometric solution that budgeted optimization cannot reliably discover, while subsequent optimization recovers scene-specific details beyond the feed-forward prior.



\textit{Amortized Anchor Refinement} therefore lets a frozen feed-forward backbone predict the geometric anchor and spends the entire optimization budget on refining it to the target scene.
A \emph{capacity floor} keeps the field dense throughout refinement, and a training-free \emph{persistence floor} protects thin structures when the field is pruned for transmission.
The fitted trajectories are streamed directly as scene flow, preserving arbitrary-time queries without per-frame optimization or residual coding.
On a single consumer GPU, our cascade reaches $\mainboth$~dB at a $\peakoff$~GB fitting peak with $5.4\times$ fewer Gaussians than a heavyweight optimizer.
Together, these results demonstrate that continuous-time 4D reconstruction can be learned, compressed, and played back within the resources of a standalone XR headset.

\noindent
Our contributions are fourfold:
\begin{itemize}[leftmargin=*,itemsep=2pt]
    \item We identify \emph{budget-constrained continuous-time 4D reconstruction} as a distinct problem rather than a scaled-down one, and characterize its two fundamental failure modes: a ceiling on what feed-forward prediction reaches, and the collapse of optimization that starts without an anchor.
    
    \item We propose \textbf{Amortized Anchor Refinement}, which combines a grounded feed-forward anchor, a mirror-symmetry prior, and a capacity-floored refinement to enable stable consumer-budget optimization without task-specific pretraining.
    
    \item We introduce \textbf{Persistent Homology Pruning and Compression}, the first Gaussian pruning method based on persistent homology, which preserves topologically important structures during compression while transmitting quantized Gaussian trajectories as a compact continuous-time scene-flow representation.
    
    \item We validate our approach on the Stage-Capture benchmark and show through controlled ablations that the improvements arise from the proposed design rather than from increased model capacity or scale.
\end{itemize}

\section{Related Work}
\label{sec::related}

\subsection{Reconstructing a Dynamic Gaussian Field}
3D Gaussian Splatting~\cite{kerbl3Dgaussians} enabled real-time novel-view synthesis, and two paradigms have since extended it to dynamic scenes.
The first deforms a canonical Gaussian representation through a learned motion field, as in Deformable-3D-Gaussians~\cite{yang2024deformable} and 4D-GS~\cite{wu20244d}.
The second assigns each Gaussian an explicit trajectory and lifetime, as in Spacetime Gaussians~\cite{li2024spacetime}, 4D-Rotor Gaussians~\cite{duan20244drotor}, Ex4DGS~\cite{lee2024ex4dgs}, and RetimeGS~\cite{retimegs}.
We adopt the latter because it supports continuous-time queries without a deformation network, and because each Gaussian trajectory directly represents the scene flow that can later be streamed without per-frame estimation or optimization.

Despite their different motion parameterizations, existing optimization-based methods share the same optimization strategy.
Each scene is reconstructed independently through lengthy optimization, typically requiring tens of minutes, hundreds of thousands of Gaussians, and workstation-class GPUs.
The deployment budget of standalone XR headsets is therefore absent from both their optimization objectives and their evaluation.
As shown in Sec.~\ref{sec::exp}, these optimization schedules collapse once that deployment budget is imposed.

An alternative direction amortizes reconstruction through feed-forward prediction.
Pose-free geometry backbones~\cite{vggt,dust3r} estimate geometry without calibrated cameras, while large reconstruction models~\cite{movies,l4gm} directly predict renderable dynamic Gaussian fields.
These methods inherit the feed-forward parameterization of static Gaussian reconstruction, where one Gaussian is predicted per input pixel~\cite{charatan2024pixelsplat,chen2024mvsplat,tang2024lgm}, limiting reconstruction to the geometric prior learned during pretraining.
Recent amortized dynamic methods~\cite{xu20254dgt} retain the same formulation while requiring large-scale task-specific pretraining.
Instead, we use a frozen geometry backbone~\cite{vggtomega} only to predict an initial geometric anchor.
Scene-specific optimization then refines this anchor, avoiding both task-specific pretraining and the quality ceiling of feed-forward prediction.

\subsection{Compacting and Delivering a Fitted Field}

Memory requirement is traditionally controlled by the densify-and-prune schedule introduced in~\cite{kerbl3Dgaussians}, while post-processing methods reduce the size of an already-optimized field through pruning, quantization, and anchor-based parameter sharing~\cite{fan2024lightgaussian,lu2024scaffold}.
These approaches incur the full optimization cost before compression begins.
In contrast, our refinement directly optimizes a compact representation by maintaining an anchored initialization above an explicit \emph{capacity floor}.
Peak memory is then dominated by the frozen backbone, whose activations are CPU-offloaded so that the entire pipeline fits within a $4$\,GB deployment budget.

Existing streaming systems focus on transmitting dynamic Gaussian fields after optimization.
3DGStream caches per-frame transforms~\cite{threedgstream}, QUEEN encodes inter-frame attribute residuals~\cite{queen}, and V$^3$ packs Gaussian attributes into videos for hardware decoding~\cite{videogs}.
These methods compress temporal changes, while the underlying Gaussian representation is still determined by pruning strategies that prioritize visible mass and often discard thin structures.
Persistent homology has recently been introduced as a training regularizer for Gaussian splatting~\cite{topologygs}, but to our knowledge has not been used as a pruning criterion.
Our streaming stage instead applies an explicit \emph{persistence floor}, complementing the capacity floor used during refinement, and directly transmits the optimized Gaussian trajectories without per-frame optimization or residual coding.
\section{Method}
\label{sec::method}

\begin{figure*}[t]
\centering
\includegraphics[width=\textwidth]{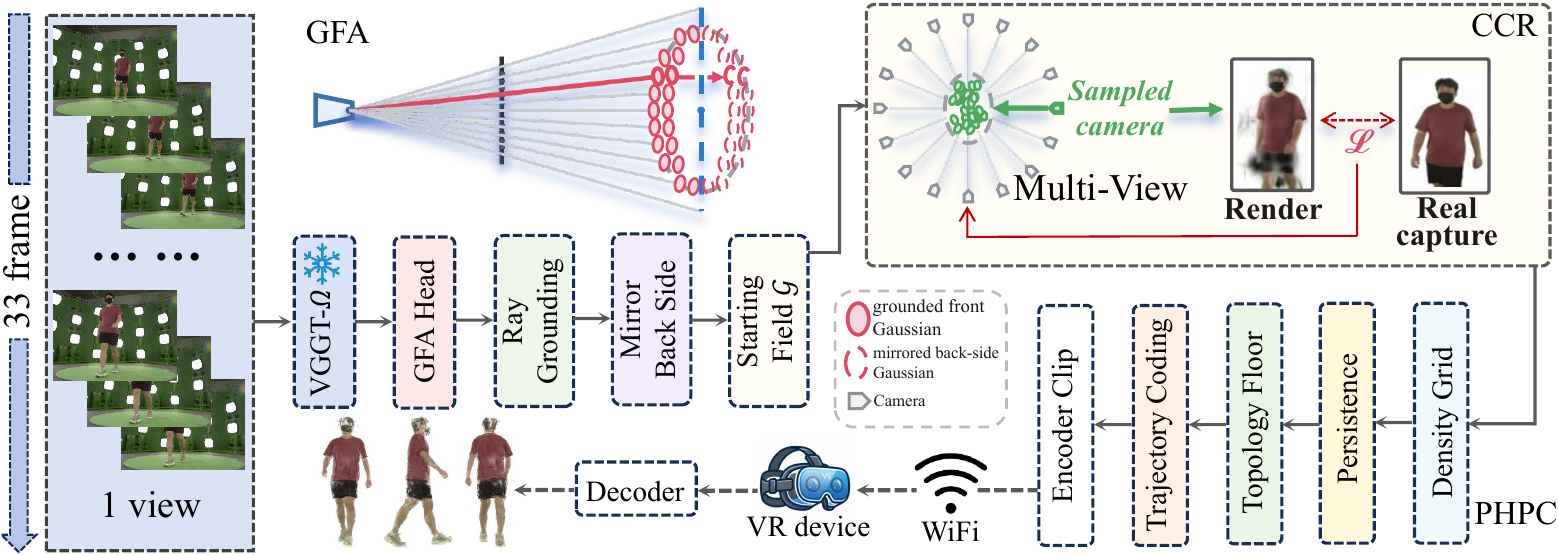}
\caption{Pipeline of \textbf{Amortized Anchor Refinement}. A grounded feed-forward anchor initializes an optimization-ready Gaussian field, which is refined under capacity constraints to recover high-quality dynamic geometry. A persistent homology pruning and compression (PHPC) stage compresses the field while preserving perceptually important structures.}
\label{fig::framework}
\end{figure*}

To make continuous-time 4D reconstruction deployable under the memory budget of a standalone XR headset, we propose the \textit{Amortized Anchor Refinement} framework, as illustrated in Fig.~\ref{fig::framework}.

\subsection{Preliminaries and Anchor Backbone}
\label{sec::pre}
Continuous-time 4D reconstruction aims to render a dynamic scene at any query viewpoint and frame in real time.
Given a set of calibrated multi-view video streams $\bm{I} = \{\bm{I}_{c,t}\}$, with $\bm{I}_{c,t} \in \mathbb{R}^{H \times W \times 3}$, where $c \in \{1, \dots, N_{c}\}$ indexes the calibrated views and $t \in [0,1]$ denotes the normalized frame, our goal is a 4D Gaussian field that renders cleanly at held-out viewpoints and frames.
Following~\cite{vggt}, we adopt a frozen pose-free VGGT-$\Omega$ as the geometry backbone.
Given the input views, the backbone encodes each view $c$ into a per-pixel depth field $\bm{D}_{c} \in \mathbb{R}^{H \times W}$ and a per-pixel feature tensor $\bm{F}_{c} \in \mathbb{R}^{H \times W \times C_{f}}$ in a single forward pass, without requiring camera poses,
\begin{equation}
\label{eq::backbone}
(\bm{D}_{c},\, \bm{F}_{c}) = \Phi(\bm{I}_{c}), \quad c \in \{1, \dots, N_{c}\},
\end{equation}
where $\Phi(\cdot)$ denotes the frozen backbone network.
The backbone is never fine-tuned and requires no task-specific pretraining. Its cost is a single forward pass.
This forward pass dominates the memory footprint of the pipeline, so we CPU-offload the backbone layer by layer at FP32, bounding peak memory by a single layer rather than by the full network.

We represent the dynamic scene as $N$ 4D Gaussians, each with a canonical center $\bm{\mu}_{i}$, a covariance $\bm{\Sigma}_{i}$, an opacity $o_{i}$, a view-dependent color $\bm{c}_{i}$, a temporal center $\tau_{i}$, and a window width $s_{i}$.
To express motion without a separate deformation residual, we adopt the short-tailed temporal-opacity design of RetimeGS~\cite{retimegs}.
Each Gaussian follows a low-order trajectory whose center moves in time as
\begin{equation}
\label{eq::traj}
\bm{\mu}_{i}(t) = \bm{\mu}_{i} + \bm{v}^{1}_{i}\,\delta_{i} + \bm{v}^{2}_{i}\,\delta_{i}^{2} + \bm{v}^{3}_{i}\,\delta_{i}^{3}, \quad \delta_{i} = t - \tau_{i},
\end{equation}
where $\bm{v}^{1}_{i}, \bm{v}^{2}_{i}, \bm{v}^{3}_{i} \in \mathbb{R}^{3}$ are the learned velocity, acceleration, and jerk terms.
Each Gaussian is visible only within a compact temporal window,
\begin{equation}
\label{eq::topacity}
o_{i}(t) = o_{i}\,\exp\!\left(-\frac{(t-\tau_{i})^{2}}{2\,s_{i}^{2}}\right),
\end{equation}
so that it contributes only near its own frame $\tau_{i}$.
At a query time $t$, the field is rasterized to view $c$ by the standard splatting composite
\begin{equation}
\label{eq::render}
\bm{C}(\bm{p}) = \sum_{i \in \mathcal{N}} \bm{c}_{i}\,\alpha_{i}(t) \prod_{j < i}\bigl(1 - \alpha_{j}(t)\bigr),
\end{equation}
where $\bm{p}$ is a pixel, $\mathcal{N}$ the depth-ordered Gaussians overlapping it, and $\alpha_{i}(t)$ combines the projected 2D footprint of Gaussian $i$ with its temporal opacity in Eq.~\ref{eq::topacity}.

\subsection{Grounded Feed-forward Anchor}
\label{sec::gfa}
The first stage, named \textit{Grounded Feed-forward Anchor (GFA)}, predicts an initialization for the per-scene optimizer.
Regressing Gaussian centers as \emph{free} 3D coordinates fails to optimize. Free coordinates have no geometric correspondence to the input observations, so the photometric gradient provides no consistent direction.
Instead, we \emph{ground} every predicted Gaussian on its input-view ray. The frozen backbone already localizes the scene to within a viewing ray, making ray grounding sufficient to remove this ambiguity while preserving local geometric flexibility.
As illustrated in the \textit{GFA} block of Fig.~\ref{fig::framework}, given the backbone depth $\bm{D}_{c}(\bm{p})$, camera center $\bm{o}_{c}$, and ray direction $\bm{d}_{c}(\bm{p})$, the anchored Gaussian center at pixel $\bm{p}$ in view $c$ is
\begin{equation}
\label{eq::ground}
\bm{\mu}
=
\bm{o}_{c}
+
\bm{D}_{c}(\bm{p})\,\bm{d}_{c}(\bm{p})
+
\Delta\bm{\mu},
\end{equation}
where $\Delta\bm{\mu}$ is a learnable residual offset from the ray.
A lightweight prediction head $g_{\theta}$ takes the backbone feature $\bm{F}_{c}(\bm{p})$ and predicts the remaining Gaussian parameters,
\begin{equation}
\label{eq::head}
\bigl(
\Delta\bm{\mu},
\bm{\Sigma},
o,
\bm{c},
\tau,
s,
\bm{v}^{1:3}
\bigr)
=
g_{\theta}\!\bigl(\bm{F}_{c}(\bm{p})\bigr),
\end{equation}
where the head emits $K$ Gaussians along each ray at increasing depths, with $K{=}2$ by default, forming a thin volumetric anchor rather than a single surface.
%

\noindent\textbf{Mirror-symmetry back-side anchor.}
The input views observe only the front surface of the subject, leaving the ray-grounded anchor of Eq.~\ref{eq::ground} hollow on the occluded side.
This missing geometry is largely corrected when dense temporal supervision is available. Under sparse-time supervision, however, the unconstrained back side allows per-Gaussian trajectories to overfit the observed frames and generalize poorly to held frames.
We therefore initialize the occluded region by reflecting the anchor Gaussians across the frontal symmetry plane, drawn as the \textit{Mirror} step in Fig.~\ref{fig::framework}, producing a coarse yet closed geometry that regularizes optimization.
Let $\bar{\bm{\mu}}$ denote the centroid of the subject Gaussians and $\bm{n}$ the unit normal of the frontal symmetry plane. The mirrored center is
\begin{equation}
\label{eq::mirror}
\bm{\mu}^{\mathrm{mir}}
=
\bm{\mu}
-
2\bigl[(\bm{\mu}-\bar{\bm{\mu}})\cdot\bm{n}\bigr]\bm{n},
\end{equation}
which requires neither additional parameters nor extra supervision.
This mirror-symmetry prior has little effect for long, densely supervised sequences, where observations eventually constrain the back side, but becomes critical for short clips.
The front rays and their mirrored counterparts together form the \emph{starting field}, the anchored initialization that the refinement stage takes over. 

\subsection{Capacity Constrained Refinement}
\label{sec::CCR}
The second stage refines the anchored field by fitting it directly to the raw multi-view capture. We name this module \textit{Capacity Constrained Refinement (CCR)}.
Under sparse real supervision, the standard densify-and-prune schedule rapidly collapses the representation to a few hundred Gaussians, limiting geometric detail and leaving isolated floaters.
We find that this behavior is caused primarily by aggressive opacity pruning, and replace it with an opacity-based \emph{capacity constraint}.
As illustrated in the \textit{CCR} block of Fig.~\ref{fig::framework}, after each pruning step the field is constrained to contain at least $N_{\min}$ Gaussians by retaining the $N_{\min}$ highest-opacity primitives,
\begin{equation}
\label{eq::floor}
\mathcal{K}
=
\operatorname*{top\text{-}}_{N_{\min}}
\bigl(\{o_i\}_{i=1}^{N}\bigr),
\end{equation}
where $\mathcal{K}$ denotes the retained index set.
%
%
To suppress floaters, we remove Gaussians that lie far from the scene medoid or whose scales exceed a preset threshold.
Because refinement adjusts a grounded initialization rather than searching for geometry, a short optimization suffices, and $N_{\min}$ directly bounds the memory footprint.

\subsection{Optimization Objective}
\label{sec::obj}

Only \textit{CCR} is optimized on a per-scene basis. \textit{GFA} consists of a single forward pass through the frozen backbone and requires neither optimization nor a teacher model.
The refinement minimizes a photometric objective over the masked multi-view observations,
\begin{equation}
\label{eq::loss}
\mathcal{L}
=
(1-\lambda)\mathcal{L}_{1}
+
\lambda\mathcal{L}_{\mathrm{D\text{-}SSIM}}
+
\gamma\mathcal{L}_{\mathrm{LPIPS}},
\end{equation}
where $\mathcal{L}_{1}$ enforces pixel-wise reconstruction, $\mathcal{L}_{\mathrm{D\text{-}SSIM}}$ structural consistency and $\mathcal{L}_{\mathrm{LPIPS}}$ perceptual quality, weighted by $\lambda$ and $\gamma$. Unless otherwise specified, we use $\lambda=0.2$ and $\gamma=0.1$.
%

\subsection{Persistent Homology Pruning and Compression}
\label{sec::tfs}
The refined field renders in real time, but its size exceeds a wireless streaming budget. We therefore introduce \textit{Persistent Homology Pruning and Compression (PHPC)} to prune the field for transmission.
The standard criterion ranks Gaussians by visibility mass,
\begin{equation}
\label{eq::mass}
w_i=o_i\,\bar{\sigma}_i,
\end{equation}
where $\bar{\sigma}_{i}$ denotes the mean scale of Gaussian $i$.
Mass ranking preserves appearance, but slender structures such as wrists and ankles carry little mass and are discarded first, so the subject topology breaks before texture quality degrades.

As illustrated in the \textit{PHPC} block of Fig.~\ref{fig::framework}, we prevent these failures using persistent homology derived from Morse theory~\cite{milnor1963morse}.
At each frame $t$ in a small reference set $\mathcal{T}$, the refined field defines a density that we accumulate on a voxel grid, the \emph{density grid} on which the diagram below is computed,
\begin{equation}
\label{eq::density}
\sigma_t(\bm{x})
=
\sum_i
o_i(t)\,
\mathcal{G}\!\bigl(\bm{x};
\bm{\mu}_i(t),
\bm{\Sigma}_i
\bigr),
\end{equation}
where $\bm{x}$ is a grid point, $\mathcal{G}$ the Gaussian kernel carrying the fixed covariance $\bm{\Sigma}_i$ of Sec.~\ref{sec::pre}, and $o_i(t)$ the temporal opacity of Eq.~\ref{eq::topacity}. Its persistence diagram~\cite{edelsbrunner2010computational} summarizes the connected components of the subject across density thresholds.
Thresholding the diagram at persistence $\varepsilon$ defines a \emph{topology floor}: every feature whose persistence exceeds $\varepsilon$ must be preserved after pruning.
Given a transmission budget of $N_s$ Gaussians, the number the clip package may carry, we solve
\begin{equation}
\label{eq::topofloor}
\begin{aligned}
\mathcal{K}_s
&=
\operatorname*{top\text{-}}_{N_s}
\bigl(\{w_i\}\bigr)\\
\text{s.t.}\quad
&
\mathrm{Dgm}_{\varepsilon}
\!\left(
\sigma_t^{\mathcal{K}_s}
\right)
=
\mathrm{Dgm}_{\varepsilon}
\!\left(
\sigma_t
\right),
\qquad
t\in\mathcal{T},
\end{aligned}
\end{equation}
where $\mathrm{Dgm}_{\varepsilon}$ denotes the persistence diagram thresholded at $\varepsilon$, and $\sigma_t^{\mathcal{K}_s}$ the density of the retained subset $\mathcal{K}_s$.
We enforce this constraint with a \emph{prune-verify-restore-reverify} procedure.
After pruning, we recompute the persistence diagram and restore the highest-ranked discarded Gaussians associated with every missing persistent feature, while removing the lowest-ranked retained Gaussians to maintain the target budget.
The restored field is accepted only if it preserves a strictly larger fraction of $\varepsilon$-persistent features; otherwise the pruned field is kept, so the floor is activated only when it is needed.
The sender also has to choose the ranking. A ranking is a heuristic for which Gaussians carry the subject, and no single heuristic holds across scenes and budgets.
We therefore prune with each candidate ranking, apply the floor to each, and keep the package that best reproduces the field it was pruned from, measured by rendering both and comparing them.
This comparison is available at the sender because it references the fitted field rather than ground-truth images, so the choice can be made at deployment time, and the transmitted package is never worse than the best individual criterion by more than the resolution of this measurement.

The temporal representation requires no additional modeling.
The trajectories of Eq.~\ref{eq::traj} already encode the scene flow, so compression reduces to efficient coding.
For each Gaussian, this \emph{trajectory coding} step quantizes the nine coefficients to $10$ bits and entropy-codes them together with the FP16/INT8-quantized static attributes.
The encoder runs once on the server; the result is a self-contained clip that the decoder on the edge client turns directly into Eq.~\ref{eq::render}, with no server interaction during rendering.
\section{Experiments}
\label{sec::exp}

\begin{table*}[t]
\centering
\caption{\textbf{Per-scene comparison.} Foreground-masked held-timestamp PSNR (dB). \textcolor{gray}{RetimeGS} exceeds the deployment budget; FILM$^\dagger$ interpolates in 2D.}
\label{tab::sota}
\begin{threeparttable}
\setlength{\tabcolsep}{3.6pt}
\begin{tabular}{l ccccccccc c}
\toprule
Method & walk & bear & doll & newund. & openc. & passd. & pickupd. & putcomp. & stretch & \textbf{Mean}\\
\midrule
FILM~\cite{film}$^\dagger$                   & 29.88 & 29.27 & 27.40 & 31.46 & 34.10 & 29.34 & 29.86 & 32.17 & 27.59 & 30.12\\
\midrule
Deformable-3DGS~\cite{yang2024deformable}    & 13.30 & 17.87 & 18.72 & 22.28 & 24.07 & 17.16 & 16.66 & 23.17 & 15.33 & 18.73\\
GaussianFlow~\cite{gaussianflow}     & 21.27 & 18.78 & 17.73 & 23.17 & 20.21 & 21.72 & 18.53 & 21.81 & 17.30 & 20.06\\
4D-GS~\cite{wu20244d}                        & 18.28 & 19.06 & 18.87 & 25.49 & 25.30 & 20.35 & 20.12 & 24.90 & 18.44 & 21.20\\
\textcolor{gray}{RetimeGS~\cite{retimegs}} & \textcolor{gray}{24.57} & \textcolor{gray}{22.09} & \textcolor{gray}{26.02} & \textcolor{gray}{29.56} & \textcolor{gray}{29.67} & \textcolor{gray}{26.44} & \textcolor{gray}{25.38} & \textcolor{gray}{29.81} & \textcolor{gray}{24.96} & \textcolor{gray}{\teachmean}\\
\midrule
\textcolor{linkblue}{Ours, short clip (9 fr.)} & \textcolor{linkblue}{22.91} & \textcolor{linkblue}{20.41} & \textcolor{linkblue}{18.84} & \textcolor{linkblue}{23.19} & \textcolor{linkblue}{23.53} & \textcolor{linkblue}{14.66} & \textcolor{linkblue}{24.93} & \textcolor{linkblue}{19.34} & \textcolor{linkblue}{25.44} & \textcolor{linkblue}{21.47}\\
\textbf{\textcolor{linkblue}{Ours, full action (33 fr.)}} & \textbf{\textcolor{linkblue}{24.22}} & \textbf{\textcolor{linkblue}{22.46}} & \textbf{\textcolor{linkblue}{22.96}} & \textbf{\textcolor{linkblue}{26.60}} & \textbf{\textcolor{linkblue}{27.96}} & \textbf{\textcolor{linkblue}{22.87}} & \textbf{\textcolor{linkblue}{23.39}} & \textbf{\textcolor{linkblue}{27.14}} & \textbf{\textcolor{linkblue}{21.19}} & \textbf{\textcolor{linkblue}{\mainmean}}\\
\bottomrule
\end{tabular}
\begin{tablenotes}
\item[$\dagger$] FILM performs 2D frame interpolation independently for each training camera by synthesizing only the temporal midpoint between two neighboring supervised frames. Because it does not reconstruct a 4D scene representation, it cannot render novel viewpoints or arbitrary frames and is therefore not a deployable 4D method.
\end{tablenotes}
\end{threeparttable}
\end{table*}

\subsection{Experimental Setup}

\noindent\textbf{Dataset.}
We evaluate on the multi-view Stage-Capture dataset released with RetimeGS~\cite{retimegs}, nine dynamic scenes of human motion and human-object interaction recorded by $32$ synchronized calibrated cameras at $1024{\times}750$ with per-frame foreground masks.
Each clip spans $33$ frames of a complete action with large inter-frame motion, non-rigid deformation, and severe self-occlusion, the regime in which temporal interpolation is hardest.

\noindent\textbf{Experiment Protocol.}
We follow the dataset's official every-other-frame hold-out split: all $32$ calibrated cameras and the $17$ even-indexed frames are used for training, and the $16$ odd-indexed frames are held out. Our primary evaluation measures continuous-time interpolation on the held-out frames with foreground-masked PSNR, SSIM and LPIPS; we also report reconstruction quality on the observed frames.
All experiments use a fixed random seed. Residual run-to-run variation due to nondeterministic CUDA atomics in \texttt{gsplat} is within $\pm0.3$~dB, consistent with standard 3DGS implementations.

\noindent\textbf{Implementation.}
The backbone is a frozen VGGT-$\Omega$~\cite{vggtomega}, never fine-tuned and CPU-offloaded at FP32.
Our \textit{GFA} module predicts $K{=}2$ grounded Gaussians per input ray.
Our \textit{CCR} optimizes the field for $96$k Adam steps at $512{\times}375$ with a capacity floor of $N_{\min}{=}2{\times}10^5$, a temporal window scale of $1.5$, and geometric floater filtering, minimizing
$0.8\,\mathcal{L}_1 + 0.2\,\mathcal{L}_{\mathrm{D\text{-}SSIM}} + 0.1\,\mathcal{L}_{\mathrm{LPIPS}}$ using exponential learning-rate decay over the final $10\%$ of training.
Densification runs over the first $70\%$ of the steps, or the first $35\%$ on scenes whose Gaussian count grows without settling.
For the nine-frame sequence, the same protocol is used with the capacity floor scaled to $N_{\min}{=}5{\times}10^4$ and the optimization shortened to $24$k steps; the mirror-symmetry anchor is enabled only in this setting.
All methods are evaluated on a single $24$~GB RTX 6000 GPU under this protocol, so quality, memory, runtime and Gaussian count are directly comparable.

\subsection{Comparison with State of the Art}
\textit{Amortized Anchor Refinement} achieves the best accuracy on every scene in Table~\ref{tab::sota}, outperforming Deformable-3DGS by an average of $+\dgsgap$~dB and by up to $+10.9$~dB on \emph{walk}.
The baselines are limited by temporal interpolation: coordinate-based MLPs overfit the observed frames and ghost between them, while flow-supervised fields and spatio-temporal grids recover part of the gap. Our anchored short-tailed trajectory is defined continuously in time and never interpolates unseen frames.
Although RetimeGS achieves higher reconstruction quality, Table~\ref{tab::sotadeploy} highlights its substantially higher deployment cost. All competing 4D Gaussian methods require $6$--$9$~GB during optimization, exceeding a $4$~GB consumer budget, whereas our method peaks at only $\peakoff$~GB while also delivering the fastest rendering speed.
FILM attains the highest held-timestamp accuracy but performs 2D video interpolation rather than reconstructing a dynamic 3D representation, so it cannot synthesize novel viewpoints or arbitrary frames; held-time PSNR alone does not characterize the task.

\begin{table}[t]
\centering
\caption{\textbf{Deployment efficiency.} \#G ($10^4$), peak optimization VRAM (GB), fitting time (min) and held-timestamp PSNR (dB). \#G and PSNR are nine-scene means; VRAM is measured on \emph{walk}.}
\label{tab::sotadeploy}
\setlength{\tabcolsep}{4pt}
\begin{tabular}{l cccc}
\toprule
Method & \#G$\downarrow$ & VRAM$\downarrow$ & Fit$\downarrow$ & PSNR$\uparrow$\\
\midrule
Deformable-3DGS~\cite{yang2024deformable} & \dgsG & \dgsvram & \dgstime & \dgsmean\\
GaussianFlow~\cite{gaussianflow}          & \gflowG & \gflowvram & \gflowtime & \gflowmean\\
4D-GS~\cite{wu20244d}                     & \fourdgsG & \fourdgsvram & \fourdgstime & \fourdgsmean\\
\textcolor{gray}{RetimeGS~\cite{retimegs}} & \textcolor{gray}{\retimegs} & \textcolor{gray}{\retimevram} & \textcolor{gray}{136} & \textcolor{gray}{\teachmean}\\
\midrule
\textcolor{linkblue}{Ours (9 fr.)} & \textcolor{linkblue}{\ninegs} & \textcolor{linkblue}{\ninevram} & \textcolor{linkblue}{\ninetime} & \textcolor{linkblue}{\ninemean}\\
\textbf{\textcolor{linkblue}{Ours (33 fr.)}} & \textbf{\textcolor{linkblue}{\ourgs}} & \textbf{\textcolor{linkblue}{\peakoff}} & \textbf{\textcolor{linkblue}{\ourtime}} & \textbf{\textcolor{linkblue}{\mainmean}}\\
\bottomrule
\end{tabular}
\end{table}

\noindent All reported PSNR values are foreground-masked; the static background follows the standard 3DGS formulation and is composited separately.

\subsection{Generalization Across Scenes}

Across the nine scenes of Table~\ref{tab::ninescene} our method reaches a mean held-timestamp PSNR of $\mainmean{\pm}\mainstd$~dB.
Since all camera views are supervised, we evaluate temporal interpolation on two disjoint camera splits, which agree to $24.31$ against $24.11$~dB, so the reported performance is insensitive to the split. Quality degrades gradually with scene complexity: the hardest scenes are those with large articulated or heavily self-occluding motion, \emph{bear} and \emph{stretch}. No scene needs its own hyperparameters.

\begin{table*}[t]
\centering
\caption{\textbf{Per-scene results.} Foreground-masked PSNR (dB). \emph{Interp.}$_{A,B}$: held timestamps on two disjoint camera splits; \emph{Recon.}: observed timestamps.}
\label{tab::ninescene}
\setlength{\tabcolsep}{5pt}
\begin{tabular}{l ccc cc cc c}
\toprule
& \multicolumn{3}{c}{PSNR$\uparrow$ (dB)} & \multicolumn{2}{c}{SSIM$\uparrow$} & \multicolumn{2}{c}{LPIPS$\downarrow$} & \#G$\downarrow$ \\
\cmidrule(lr){2-4}\cmidrule(lr){5-6}\cmidrule(lr){7-8}
Scene & Interp.$_A$ & Recon. & Interp.$_B$ & Interp. & Recon. & Interp. & Recon. & ($10^4$) \\
\midrule
walk         & 24.22 & 26.54 & 24.03 & 0.982 & 0.990 & 0.018 & 0.010 & 22.6 \\
bear         & 22.46 & 26.43 & 22.93 & 0.968 & 0.984 & 0.033 & 0.017 & 22.6 \\
doll         & 22.96 & 26.36 & 22.75 & 0.972 & 0.984 & 0.024 & 0.012 & 22.6 \\
newundress   & 26.60 & 27.62 & 26.09 & 0.971 & 0.982 & 0.027 & 0.018 & 22.5 \\
opencase     & 27.96 & 28.30 & 27.54 & 0.982 & 0.986 & 0.017 & 0.011 & 22.4 \\
passdoll     & 22.87 & 26.87 & 22.94 & 0.945 & 0.973 & 0.055 & 0.023 & 34.9 \\
pickupdoll   & 23.39 & 26.70 & 23.41 & 0.964 & 0.980 & 0.042 & 0.021 & 22.3 \\
putcomputer  & 27.14 & 27.89 & 26.63 & 0.986 & 0.990 & 0.012 & 0.009 & 22.6 \\
stretch      & 21.19 & 26.30 & 20.66 & 0.969 & 0.985 & 0.030 & 0.014 & 22.6 \\
\midrule
\textbf{\textcolor{linkblue}{Mean}} & \textbf{\textcolor{linkblue}{24.31}} & \textbf{\textcolor{linkblue}{27.00}} & \textbf{\textcolor{linkblue}{24.11}} & \textbf{\textcolor{linkblue}{0.971}} & \textbf{\textcolor{linkblue}{0.984}} & \textbf{\textcolor{linkblue}{0.029}} & \textbf{\textcolor{linkblue}{0.015}} & \textbf{\textcolor{linkblue}{23.9}} \\
\textcolor{gray}{Std}          & \textcolor{gray}{2.22} & \textcolor{gray}{0.70} & \textcolor{gray}{2.08} & \textcolor{gray}{.012} & \textcolor{gray}{.005} & \textcolor{gray}{.013} & \textcolor{gray}{.005} & \textcolor{gray}{3.9} \\
\bottomrule
\end{tabular}
\end{table*}

\begin{figure*}[t]
\centering
\includegraphics[width=\textwidth]{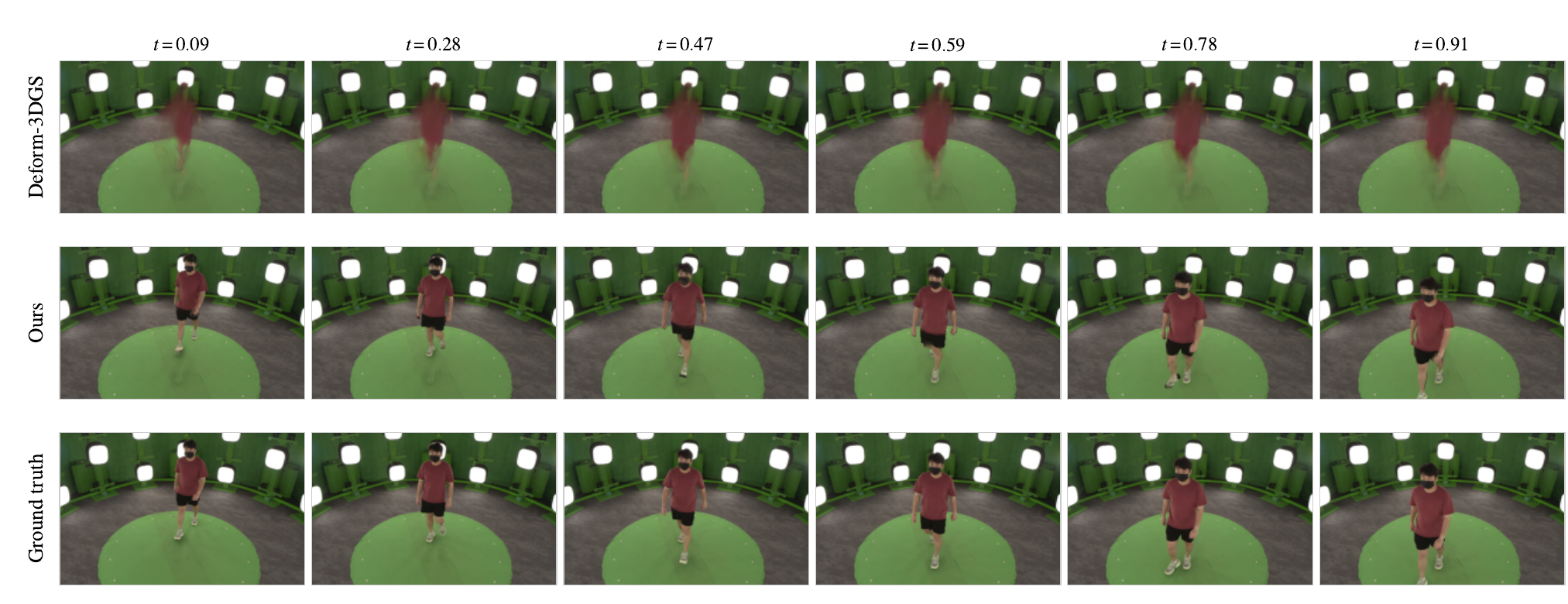}
\caption{\textbf{Held frames on a held \emph{walk} camera.} No method was supervised on these frames. Foreground composited over a static plate.}
\label{fig::qualitative}
\vspace{-1em}
\end{figure*}

\begin{figure}[t]
\centering
\includegraphics[width=\columnwidth]{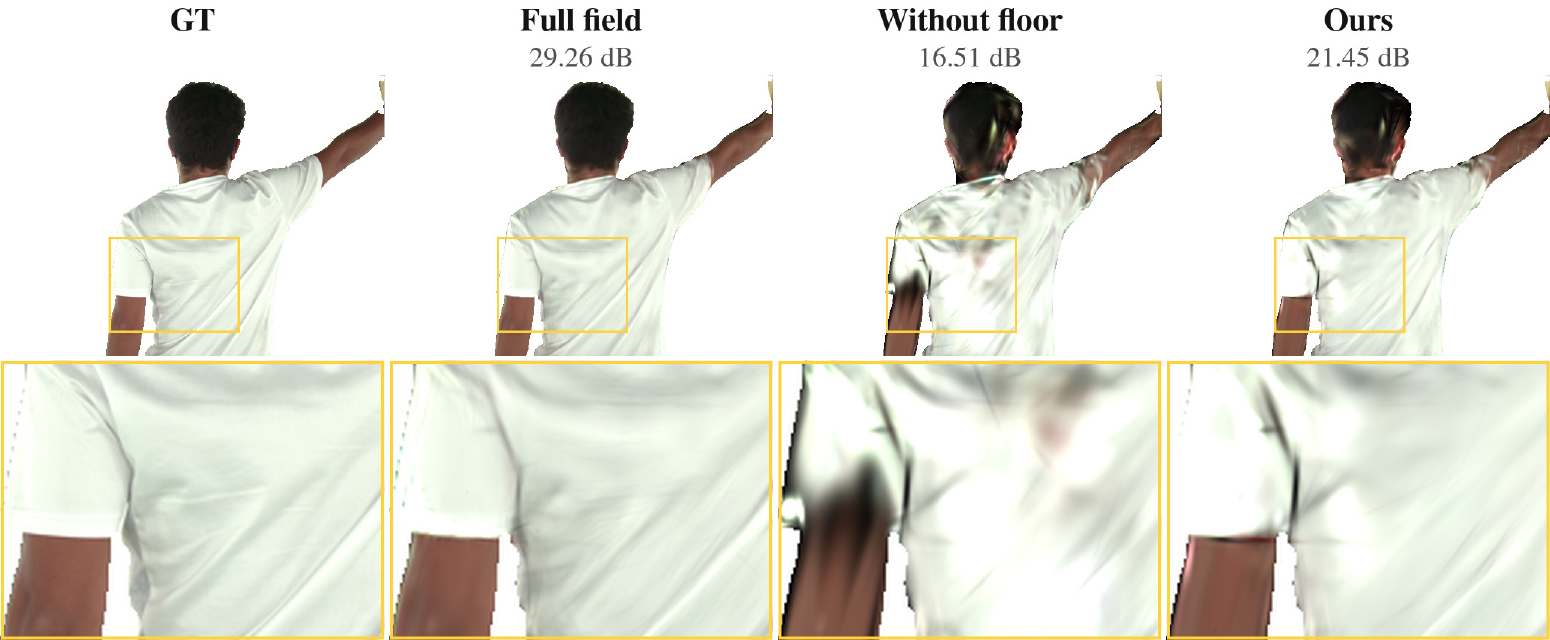}
\caption{\textbf{The verified topology floor} on \emph{doll} at $2{\times}10^{4}$ Gaussians, held camera and timestamp. Labels report PSNR at this view. Bottom: zoomed view.}
\label{fig::pruneabl}
\vspace{-1em}
\end{figure}

\subsection{Ablation Study}
\noindent\textbf{Anchor and capacity floor.}
Table~\ref{tab::ablation} adds one module at a time under an identical recipe. From-scratch optimization reaches $\ablscratch$~dB on \emph{walk}: the photometric gradient gives unanchored Gaussians no consistent direction, so the densify-and-prune schedule strips the field to $40$ Gaussians and the run ends with almost nothing to render. The same recipe on \emph{opencase} ends at $\scratchopencase$~dB with $\scratchopencaseN$ Gaussians, so the collapse is a property of the regime rather than of one scene. Anchoring the field removes it, lifting quality to $\ablgfa$~dB, but the anchored field is still pruned down to a few hundred Gaussians, because the schedule that caused the collapse is unchanged. The capacity floor addresses precisely that: holding at least $N_{\min}$ Gaussians throughout raises quality to $\mainboth$~dB at $\ourgs{\times}10^{4}$ Gaussians. Neither module is sufficient alone: the anchor supplies the geometry the optimizer cannot find, and the floor keeps enough capacity for it to be refined.

\noindent\textbf{Mirror-symmetry back side.}
The input views observe only the front surface, so the anchor is hollow behind the subject. Table~\ref{tab::ablation} evaluates the mirrored completion in the nine-frame regime, where temporal supervision is sparse: removing it drops $\ninemirror$ to $\ninenomirror$~dB. The optimization curve explains the size of the gap. Without the back side, quality peaks at $\ninenomirrorpeak$~dB after a few thousand steps and then decays monotonically: the unconstrained rear Gaussians move into the observed frames and the trajectories overfit them, so the final number is the end of a decay rather than a plateau. With the mirrored anchor the same run improves throughout. The prior leaves the full-length sequence unchanged, where dense supervision constrains the rear surface on its own, so we enable it only for short clips.

\noindent\textbf{Topology floor.}
The middle pair of Table~\ref{tab::tfs} isolates the floor on a fixed visibility-mass ranking. Above $4{\times}10^{4}$ Gaussians nothing $\varepsilon$-persistent dies, so the floor never activates and the two rows are identical. At $2{\times}10^{4}$ the ranking starts severing thin structures: preservation falls to $0.82$, and restoring the Gaussians that carry the missing features returns it to $0.88$ and raises decoded quality from $19.52$ to $20.30$~dB. The floor is insurance rather than a booster, inert when the ranking is already safe and recovering what a ranking destroys once the budget tightens, as Figure~\ref{fig::pruneabl} shows at the sleeve boundary.

\begin{table}[t]
\centering
\caption{\textbf{Module ablation} on \emph{walk} under an identical recipe within each block. PSNR: foreground-masked held-timestamp PSNR (dB); \#G: Gaussians after optimization.}
\label{tab::ablation}
\setlength{\tabcolsep}{4pt}
\begin{tabular}{@{}l cc cc@{}}
\toprule
& GFA & CCR & PSNR$\uparrow$ & \#G ($10^4$)\\
\midrule
\multicolumn{5}{@{}l}{$33$-frame sequence}\\
From scratch      &        &        & \ablscratch & $0.004$\\
$+$ anchor        & \cmark &        & \ablgfa     & $0.024$\\
\textbf{\textcolor{linkblue}{$+$ capacity floor}} & \textbf{\textcolor{linkblue}{\cmark}} & \textbf{\textcolor{linkblue}{\cmark}} & \textbf{\textcolor{linkblue}{\mainboth}} & \textbf{\textcolor{linkblue}{\ourgs}}\\
From scratch, \emph{opencase} & & & \scratchopencase & $0.006$\\
\midrule
\multicolumn{5}{@{}l}{$9$-frame clip}\\
Anchor without mirror & \cmark & \cmark & \ninenomirror & $5.5$\\
\textbf{\textcolor{linkblue}{Anchor with mirror}} & \textbf{\textcolor{linkblue}{\cmark}} & \textbf{\textcolor{linkblue}{\cmark}} & \textbf{\textcolor{linkblue}{\ninemirror}} & \textbf{\textcolor{linkblue}{$10.1$}}\\
\bottomrule
\end{tabular}
\end{table}

\subsection{Analysis}

\noindent\textbf{Refinement budget and clip length.}
Quality rises with optimization length but is insensitive to the Gaussian budget once the capacity floor is reached.
A nine-frame clip is anchored and refined in $\ninetime$~min and the full sequence in $\ourtime$~min, with the shorter clip varying more because fewer observations constrain it.

\noindent\textbf{Deployment on a 4\,GB budget.}
Peak memory is dominated by the backbone forward pass rather than refinement, whose peak is only $\peakopt$~GB; CPU offloading the frozen backbone at FP32 brings the end-to-end footprint to $\peakoff$~GB without affecting quality, and the resulting representation renders at $\ourfps$~fps.

\noindent\textbf{Interpolation versus reconstruction.}
All results above hold out alternate frames; with every captured frame available the field reaches $\demoboth$~dB. Figure~\ref{fig::qualitative} shows the difference: deformation-field methods interpolate only between fitted observations, whereas our trajectories are continuous in time.

\begin{table}[t]
\centering
\caption{\textbf{Streaming quality per budget.} Nine-scene means. Mbps: coded bitrate at $25$~fps; PSNR (dB): after decoding; Topo: fraction of $\varepsilon$-persistent features preserved. The middle pair isolates the topology floor on a fixed ranking.}
\label{tab::tfs}
\setlength{\tabcolsep}{4pt}
\begin{tabular}{l ccccc}
\toprule
$N_{s}$ ($10^4$) & 8.0 & 4.0 & 2.0 & 1.0 & 0.5\\
Mbps & 33.4 & 16.8 & \phantom{0}8.4 & \phantom{0}4.2 & \phantom{0}2.1\\
\midrule
\multicolumn{6}{@{}l}{PSNR$\uparrow$}\\
LightGaussian~\cite{fan2024lightgaussian} & 24.00 & 23.12 & 19.88 & 16.83 & 15.27\\
RadSplat~\cite{niemeyer2024radsplat}  & 24.00 & 23.66 & 21.90 & 17.98 & 13.17\\
Visibility mass & 24.31 & 23.66 & 19.52 & 16.57 & 14.81\\
\quad $+$ floor & 24.31 & 23.62 & 20.30 & 16.79 & 14.85\\
\textbf{\textcolor{linkblue}{Ours}} & \textbf{\textcolor{linkblue}{24.31}} & \textbf{\textcolor{linkblue}{24.28}} & \textbf{\textcolor{linkblue}{22.59}} & \textbf{\textcolor{linkblue}{19.70}} & \textbf{\textcolor{linkblue}{16.71}}\\
\midrule
\multicolumn{6}{@{}l}{Topo$\uparrow$}\\
LightGaussian~\cite{fan2024lightgaussian} & 1.00 & 0.98 & 0.85 & 0.70 & 0.60\\
RadSplat~\cite{niemeyer2024radsplat}  & 1.00 & 1.00 & 0.95 & 0.80 & 0.62\\
Visibility mass & 1.00 & 0.98 & 0.82 & 0.69 & 0.58\\
\quad $+$ floor & 1.00 & 0.99 & 0.88 & 0.72 & 0.60\\
\textbf{\textcolor{linkblue}{Ours}} & \textbf{\textcolor{linkblue}{1.00}} & \textbf{\textcolor{linkblue}{1.00}} & \textbf{\textcolor{linkblue}{0.98}} & \textbf{\textcolor{linkblue}{0.90}} & \textbf{\textcolor{linkblue}{0.79}}\\
\bottomrule
\end{tabular}
\end{table}

\subsection{Streaming to Edge Devices}
\label{sec::tfsexp}
We evaluate \emph{Persistent Homology Pruning and Compression} on the fitted fields of all nine scenes against two training-free criteria from current Gaussian compression systems: the global significance score of LightGaussian~\cite{fan2024lightgaussian} and the accumulated blending contribution of RadSplat~\cite{niemeyer2024radsplat}.
All criteria rank the same field and share the same coding stage, so the comparison isolates which Gaussians are kept. We also report the \emph{topology preservation rate}, the fraction of $\varepsilon$-persistent features preserved, which quantifies Eq.~\ref{eq::topofloor}.

Our packages are the strongest at every budget below the point where pruning becomes lossless, and the margin widens as the budget tightens: both baselines weight each Gaussian by a footprint term and remove thin structures first.
No single ranking is best everywhere: at the tighter budgets opacity wins on most scenes, the blending contribution and visibility mass each win on one, and on one scene opacity collapses. Selecting the package by its fidelity to the unpruned field recovers the best candidate on almost every scene-budget pair, which places our rows above every criterion.

\begin{figure}[t]
\centering
\includegraphics[width=\columnwidth]{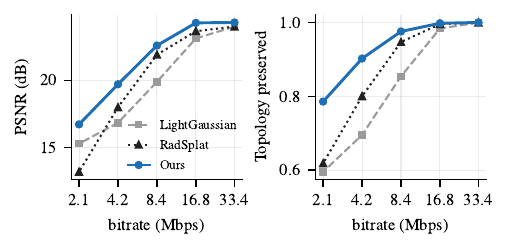}
\caption{\textbf{Streaming rate-distortion.} Nine-scene means against coded bitrate. Left: PSNR after decoding. Right: fraction of $\varepsilon$-persistent features preserved.}
\label{fig::tfsrd}
\end{figure}

\noindent\textbf{Clip package and bitrate.}
The raw FP32 representation averages $\rawmb$~MB per clip; quantizing the static attributes to FP16/INT8 and the nine trajectory coefficients to $10$ bits costs less than $0.01$~dB. The coded packages are $\tfsfortymb$ and $\tfstwentymb$~MB at $4{\times}10^{4}$ and $2{\times}10^{4}$ Gaussians, which stream at $\tfsfortymbps$ and $\tfstwentymbps$~Mbps and render at $\fpsforty$ and $\fpstwenty$~fps.

\noindent\textbf{Deployment on an untethered headset.}
On a Meta Quest Pro a WebGL viewer decodes the representation in shader: the $4{\times}10^{4}$-Gaussian package transfers over WiFi in $\questxfer$~s at $\questmbps$~Mbps against $\questxferfull$~s for the uncompressed field, and renders in immersive stereo at $\questfps$~fps against the headset's $90$~Hz display. Figure~\ref{fig::tfsrd} plots the comparison across the whole bitrate range. The pipeline therefore runs end to end: a single consumer GPU, a few-megabyte package, ordinary WiFi, and interactive rendering on a standalone headset.

\section{Conclusion}
\label{sec::conclusion}
We study dynamic Gaussian reconstruction under a joint reconstruction, rendering, and hardware budget for XR deployment. Our experiments show that optimization from scratch cannot satisfy the target budget, while increasing the capacity of feed-forward prediction does not eliminate its reconstruction-quality gap, motivating feed-forward initialization followed by lightweight refinement. \textit{Amortized Anchor Refinement} implements this strategy with a frozen backbone, budget-constrained anchor refinement, and topologically constrained compression, and our experiments demonstrate that it achieves the required reconstruction quality within the target budget, enabling reconstruction on a single consumer GPU and playback on a standalone headset.

\section*{Acknowledgments}
This work was partially supported by the U.S.\ National Science Foundation under Grant No.~2316400.
This work was also partially supported by the Wallenberg AI, Autonomous Systems and Software Program (WASP) funded by the Knut and Alice Wallenberg Foundation.
The computations and data handling were enabled by the Arrhenius HPC resource provided by National Academic Infrastructure for Supercomputing in Sweden. The authors also acknowledge the South Dakota State University High Performance Computing (SDSU HPC) team for providing access to and support for the university's high-performance computing resources.

{
    \small
    \bibliographystyle{ieeenat_fullname}
    \bibliography{reference}

@article{kerbl3Dgaussians,
  title={3D Gaussian Splatting for Real-Time Radiance Field Rendering},
  author={Kerbl, Bernhard and Kopanas, Georgios and Leimk{\"u}hler, Thomas and Drettakis, George},
  journal={ACM Transactions on Graphics},
  volume={42},
  number={4},
  year={2023},
  publisher={ACM}
}

@inproceedings{wu20244d,
  title={4D Gaussian Splatting for Real-Time Dynamic Scene Rendering},
  author={Wu, Guanjun and Yi, Taoran and Fang, Jiemin and Xie, Lingxi and Zhang, Xiaopeng and Wei, Wei and Liu, Wenyu and Tian, Qi and Wang, Xinggang},
  booktitle={Proceedings of the IEEE/CVF Conference on Computer Vision and Pattern Recognition (CVPR)},
  year={2024}
}

@inproceedings{yang2024deformable,
  title={Deformable 3D Gaussians for High-Fidelity Monocular Dynamic Scene Reconstruction},
  author={Yang, Ziyi and Gao, Xinyu and Zhou, Wen and Jiao, Shaohui and Zhang, Yuqing and Jin, Xiaogang},
  booktitle={Proceedings of the IEEE/CVF Conference on Computer Vision and Pattern Recognition (CVPR)},
  year={2024}
}

@inproceedings{retimegs,
  title={RetimeGS: Continuous-Time Reconstruction of 4D Gaussian Splatting},
  author={Wang, Xuezhen and Ma, Li and Shen, Yulin and Wang, Zeyu and Sander, Pedro V.},
  booktitle={Proceedings of the IEEE/CVF Conference on Computer Vision and Pattern Recognition (CVPR)},
  pages={7340--7350},
  year={2026}
}

@inproceedings{vggt,
  title={VGGT: Visual Geometry Grounded Transformer},
  author={Wang, Jianyuan and Chen, Minghao and Karaev, Nikita and Vedaldi, Andrea and Rupprecht, Christian and Novotny, David},
  booktitle={Proceedings of the IEEE/CVF Conference on Computer Vision and Pattern Recognition (CVPR)},
  year={2025}
}

@inproceedings{vggtomega,
  title={{VGGT-$\Omega$}},
  author={Wang, Jianyuan and Chen, Minghao and Zhang, Shangzhan and Karaev, Nikita and Sch{\"o}nberger, Johannes and Labatut, Patrick and Bojanowski, Piotr and Novotny, David and Vedaldi, Andrea and Rupprecht, Christian},
  booktitle={Proceedings of the IEEE/CVF Conference on Computer Vision and Pattern Recognition (CVPR)},
  year={2026}
}

@inproceedings{dust3r,
  title={DUSt3R: Geometric 3D Vision Made Easy},
  author={Wang, Shuzhe and Leroy, Vincent and Cabon, Yohann and Chidlovskii, Boris and Revaud, J{\'e}r{\^o}me},
  booktitle={Proceedings of the IEEE/CVF Conference on Computer Vision and Pattern Recognition (CVPR)},
  year={2024}
}

@article{movies,
  title={MoVieS: Motion-Aware 4D Dynamic View Synthesis in One Second},
  author={Lin, Chenguo and Lin, Yuchen and Pan, Panwang and Yu, Yifan and Hu, Tao and Yan, Honglei and Fragkiadaki, Katerina and Mu, Yadong},
  journal={arXiv preprint arXiv:2507.10065},
  year={2025}
}

@inproceedings{l4gm,
  title={L4GM: Large 4D Gaussian Reconstruction Model},
  author={Ren, Jiawei and Xie, Kevin and Mirzaei, Ashkan and Liang, Hanxue and Zeng, Xiaohui and Kreis, Karsten and Liu, Ziwei and Torralba, Antonio and Fidler, Sanja and Kim, Seung Wook and Ling, Huan},
  booktitle={Advances in Neural Information Processing Systems (NeurIPS)},
  year={2024}
}

@inproceedings{fan2024lightgaussian,
  title={LightGaussian: Unbounded 3D Gaussian Compression with 15x Reduction and 200+ FPS},
  author={Fan, Zhiwen and Wang, Kevin and Wen, Kairun and Zhu, Zehao and Xu, Dejia and Wang, Zhangyang},
  booktitle={Advances in Neural Information Processing Systems (NeurIPS)},
  year={2024}
}

@inproceedings{lu2024scaffold,
  title={Scaffold-GS: Structured 3D Gaussians for View-Adaptive Rendering},
  author={Lu, Tao and Yu, Mulin and Xu, Linning and Xiangli, Yuanbo and Wang, Limin and Lin, Dahua and Dai, Bo},
  booktitle={Proceedings of the IEEE/CVF Conference on Computer Vision and Pattern Recognition (CVPR)},
  year={2024}
}

@inproceedings{dnarendering,
  title={DNA-Rendering: A Diverse Neural Actor Repository for High-Fidelity Human-centric Rendering},
  author={Cheng, Wei and Chen, Ruixiang and Fan, Siming and Yin, Wanqi and Chen, Keyu and Cai, Zhongang and Wang, Jingbo and Gao, Yang and Yu, Zhengming and Lin, Zhengyu and others},
  booktitle={Proceedings of the IEEE/CVF International Conference on Computer Vision (ICCV)},
  year={2023}
}

@inproceedings{film,
  title={FILM: Frame Interpolation for Large Motion},
  author={Reda, Fitsum and Kontkanen, Janne and Tabellion, Eric and Sun, Deqing and Pantofaru, Caroline and Curless, Brian},
  booktitle={Proceedings of the European Conference on Computer Vision (ECCV)},
  year={2022}
}

@article{gaussianflow,
  title={GaussianFlow: Splatting Gaussian Dynamics for 4D Content Creation},
  author={Gao, Quankai and Xu, Qiangeng and Cao, Zhe and Mildenhall, Ben and Ma, Wenchao and Chen, Le and Tang, Danhang and Neumann, Ulrich},
  journal={arXiv preprint arXiv:2403.12365},
  year={2024}
}

@inproceedings{li2024spacetime,
  title={Spacetime Gaussian Feature Splatting for Real-Time Dynamic View Synthesis},
  author={Li, Zhan and Chen, Zhang and Li, Zhong and Xu, Yi},
  booktitle={Proceedings of the IEEE/CVF Conference on Computer Vision and Pattern Recognition (CVPR)},
  year={2024}
}

@inproceedings{duan20244drotor,
  title={4D-Rotor Gaussian Splatting: Towards Efficient Novel View Synthesis for Dynamic Scenes},
  author={Duan, Yuanxing and Wei, Fangyin and Dai, Qiyu and He, Yuhang and Chen, Wenzheng and Chen, Baoquan},
  booktitle={ACM SIGGRAPH Conference Papers},
  year={2024}
}

@inproceedings{lee2024ex4dgs,
  title={Fully Explicit Dynamic Gaussian Splatting},
  author={Lee, Junoh and Won, Changyeon and Jung, Hyunji and Bae, Inhwan and Jeon, Hae-Gon},
  booktitle={Advances in Neural Information Processing Systems (NeurIPS)},
  year={2024}
}

@inproceedings{queen,
  title={QUEEN: QUantized Efficient ENcoding of Dynamic Gaussians for Streaming Free-viewpoint Videos},
  author={Girish, Sharath and Li, Tianye and Mazumdar, Amrita and Shrivastava, Abhinav and Luebke, David and De Mello, Shalini},
  booktitle={Advances in Neural Information Processing Systems (NeurIPS)},
  year={2024}
}

@inproceedings{threedgstream,
  title={3DGStream: On-the-Fly Training of 3D Gaussians for Efficient Streaming of Photo-Realistic Free-Viewpoint Videos},
  author={Sun, Jiakai and Jiao, Han and Li, Guangyuan and Zhang, Zhanjie and Zhao, Lei and Xing, Wei},
  booktitle={Proceedings of the IEEE/CVF Conference on Computer Vision and Pattern Recognition (CVPR)},
  pages={20675--20685},
  year={2024}
}

@article{videogs,
  title={V\textsuperscript{3}: Viewing Volumetric Videos on Mobiles via Streamable 2D Dynamic Gaussians},
  author={Wang, Penghao and Zhang, Zhirui and Wang, Liao and Yao, Kaixin and Xie, Siyuan and Yu, Jingyi and Wu, Minye and Xu, Lan},
  journal={ACM Transactions on Graphics},
  volume={43},
  number={6},
  year={2024},
  publisher={ACM}
}

@inproceedings{topologygs,
  title={Topology-Aware 3D Gaussian Splatting: Leveraging Persistent Homology for Optimized Structural Integrity},
  author={Shen, Tianqi and Liu, Shaohua and Feng, Jiaqi and Ma, Ziye and An, Ning},
  booktitle={Proceedings of the AAAI Conference on Artificial Intelligence},
  year={2025}
}

@book{milnor1963morse,
  title={Morse Theory},
  author={Milnor, John},
  publisher={Princeton University Press},
  year={1963}
}

@book{edelsbrunner2010computational,
  title={Computational Topology: An Introduction},
  author={Edelsbrunner, Herbert and Harer, John},
  publisher={American Mathematical Society},
  year={2010}
}

@inproceedings{charatan2024pixelsplat,
  title={pixelSplat: 3D Gaussian Splats from Image Pairs for Scalable Generalizable 3D Reconstruction},
  author={Charatan, David and Li, Sizhe Lester and Tagliasacchi, Andrea and Sitzmann, Vincent},
  booktitle={Proceedings of the IEEE/CVF Conference on Computer Vision and Pattern Recognition (CVPR)},
  year={2024}
}

@inproceedings{chen2024mvsplat,
  title={MVSplat: Efficient 3D Gaussian Splatting from Sparse Multi-view Images},
  author={Chen, Yuedong and Xu, Haofei and Zheng, Chuanxia and Zhuang, Bohan and Pollefeys, Marc and Geiger, Andreas and Cham, Tat-Jen and Cai, Jianfei},
  booktitle={European Conference on Computer Vision (ECCV)},
  year={2024}
}

@inproceedings{tang2024lgm,
  title={LGM: Large Multi-view Gaussian Model for High-Resolution 3D Content Creation},
  author={Tang, Jiaxiang and Chen, Zhaoxi and Chen, Xiaokang and Wang, Tengfei and Zeng, Gang and Liu, Ziwei},
  booktitle={European Conference on Computer Vision (ECCV)},
  year={2024}
}

@inproceedings{xu20254dgt,
  title={4DGT: Learning a 4D Gaussian Transformer Using Real-World Monocular Videos},
  author={Xu, Zhen and Li, Zhengqin and Dong, Zhao and Zhou, Xiaowei and Newcombe, Richard A. and Lv, Zhaoyang},
  booktitle={Advances in Neural Information Processing Systems (NeurIPS)},
  year={2025}
}

@inproceedings{mildenhall2020nerf,
  title={NeRF: Representing Scenes as Neural Radiance Fields for View Synthesis},
  author={Mildenhall, Ben and Srinivasan, Pratul P. and Tancik, Matthew and Barron, Jonathan T. and Ramamoorthi, Ravi and Ng, Ren},
  booktitle={European Conference on Computer Vision (ECCV)},
  year={2020}
}

@article{muller2022instantngp,
  title={Instant Neural Graphics Primitives with a Multiresolution Hash Encoding},
  author={M{\"u}ller, Thomas and Evans, Alex and Schied, Christoph and Keller, Alexander},
  journal={ACM Transactions on Graphics},
  volume={41}, number={4}, pages={102:1--102:15},
  year={2022}
}

@inproceedings{barron2022mipnerf360,
  title={Mip-NeRF 360: Unbounded Anti-Aliased Neural Radiance Fields},
  author={Barron, Jonathan T. and Mildenhall, Ben and Verbin, Dor and Srinivasan, Pratul P. and Hedman, Peter},
  booktitle={Proceedings of the IEEE/CVF Conference on Computer Vision and Pattern Recognition (CVPR)},
  year={2022}
}

@inproceedings{pumarola2021dnerf,
  title={D-NeRF: Neural Radiance Fields for Dynamic Scenes},
  author={Pumarola, Albert and Corona, Enric and Pons-Moll, Gerard and Moreno-Noguer, Francesc},
  booktitle={Proceedings of the IEEE/CVF Conference on Computer Vision and Pattern Recognition (CVPR)},
  year={2021}
}

@inproceedings{park2021nerfies,
  title={Nerfies: Deformable Neural Radiance Fields},
  author={Park, Keunhong and Sinha, Utkarsh and Barron, Jonathan T. and Bouaziz, Sofien and Goldman, Dan B. and Seitz, Steven M. and Martin-Brualla, Ricardo},
  booktitle={Proceedings of the IEEE/CVF International Conference on Computer Vision (ICCV)},
  year={2021}
}

@article{park2021hypernerf,
  title={HyperNeRF: A Higher-Dimensional Representation for Topologically Varying Neural Radiance Fields},
  author={Park, Keunhong and Sinha, Utkarsh and Hedman, Peter and Barron, Jonathan T. and Bouaziz, Sofien and Goldman, Dan B. and Martin-Brualla, Ricardo and Seitz, Steven M.},
  journal={ACM Transactions on Graphics},
  volume={40}, number={6},
  year={2021}
}

@inproceedings{fridovichkeil2023kplanes,
  title={K-Planes: Explicit Radiance Fields in Space, Time, and Appearance},
  author={Fridovich-Keil, Sara and Meanti, Giacomo and Warburg, Frederik Rahb{\ae}k and Recht, Benjamin and Kanazawa, Angjoo},
  booktitle={Proceedings of the IEEE/CVF Conference on Computer Vision and Pattern Recognition (CVPR)},
  year={2023}
}

@inproceedings{luiten2024dynamic,
  title={Dynamic 3D Gaussians: Tracking by Persistent Dynamic View Synthesis},
  author={Luiten, Jonathon and Kopanas, Georgios and Leibe, Bastian and Ramanan, Deva},
  booktitle={International Conference on 3D Vision (3DV)},
  year={2024}
}

@inproceedings{huang2024scgs,
  title={SC-GS: Sparse-Controlled Gaussian Splatting for Editable Dynamic Scenes},
  author={Huang, Yi-Hua and Sun, Yang-Tian and Yang, Ziyi and Lyu, Xiaoyang and Cao, Yan-Pei and Qi, Xiaojuan},
  booktitle={Proceedings of the IEEE/CVF Conference on Computer Vision and Pattern Recognition (CVPR)},
  year={2024}
}

@inproceedings{yang2024realtime4dgs,
  title={Real-time Photorealistic Dynamic Scene Representation and Rendering with 4D Gaussian Splatting},
  author={Yang, Zeyu and Yang, Hongye and Pan, Zijie and Zhang, Li},
  booktitle={International Conference on Learning Representations (ICLR)},
  year={2024}
}

@inproceedings{lee2024compact3dgs,
  title={Compact 3D Gaussian Representation for Radiance Field},
  author={Lee, Joo Chan and Rho, Daniel and Sun, Xiangyu and Ko, Jong Hwan and Park, Eunbyung},
  booktitle={Proceedings of the IEEE/CVF Conference on Computer Vision and Pattern Recognition (CVPR)},
  year={2024}
}

@article{broxton2020immersive,
  title={Immersive Light Field Video with a Layered Mesh Representation},
  author={Broxton, Michael and Flynn, John and Overbeck, Ryan S. and Erickson, Daniel and Hedman, Peter and DuVall, Matthew and Dourgarian, Jason and Busch, Jay and Whalen, Matt and Debevec, Paul},
  journal={ACM Transactions on Graphics},
  volume={39}, number={4},
  year={2020}
}

@article{da3,
  title={Depth Anything 3: Recovering the Visual Space from Any Views},
  author={Lin, Haotong and Chen, Sili and Liew, Junhao and Chen, Donny Y. and Li, Zhenyu and Shi, Guang and Feng, Jiashi and Kang, Bingyi},
  journal={arXiv preprint arXiv:2511.10647},
  year={2025}
}

@article{pi3,
  title={$\pi^3$: Scalable Permutation-Equivariant Visual Geometry Learning},
  author={Wang, Yifan and Zhou, Jianjun and Zhu, Haoyi and Chang, Wenzheng and Zhou, Yang and Li, Zizun and Chen, Junyi and Pang, Jiangmiao and Shen, Chunhua and He, Tong},
  journal={arXiv preprint arXiv:2507.13347},
  year={2025}
}

@article{mapanything,
  title={MapAnything: Universal Feed-Forward Metric 3D Reconstruction},
  author={Keetha, Nikhil Varma and M{\"u}ller, Norman and Sch{\"o}nberger, Johannes and Porzi, Lorenzo and Zhang, Yuchen and Fischer, Tobias and Knapitsch, Arno and Zauss, Duncan and Weber, Ethan and Antunes, Nelson and Luiten, Jonathon and Lopez-Antequera, Manuel and Rota Bul{\`o}, Samuel and Richardt, Christian and Ramanan, Deva and Scherer, Sebastian and Kontschieder, Peter},
  journal={arXiv preprint arXiv:2509.13414},
  year={2025}
}

@inproceedings{niemeyer2024radsplat,
  author    = {Michael Niemeyer and Fabian Manhardt and Marie{-}Julie Rakotosaona and Michael Oechsle and Daniel Duckworth and Rama Gosula and Keisuke Tateno and John Bates and Dominik Kaeser and Federico Tombari},
  title     = {RadSplat: Radiance Field-Informed Gaussian Splatting for Robust Real-Time Rendering with 900+ {FPS}},
  booktitle = {International Conference on 3D Vision (3DV)},
  year      = {2025}
}
}

\clearpage
\appendix
\section{Choice of Feed-forward Backbone}
\label{sec::appbackbone}

The Grounded Feed-forward Anchor uses a frozen backbone to initialize the Gaussian field.
This backbone can in principle be replaced by any network that predicts per-view geometry.
We evaluate five alternatives to quantify how sensitive the pipeline is to this choice.

\noindent\textbf{Protocol.}
Each backbone provides a single output: a depth map per input view.
All subsequent stages are held fixed, including ray grounding, mirror-symmetry completion, attribute initialization, refinement, and evaluation.
No backbone-specific head is trained, and Gaussian colors are sampled from the input views rather than predicted, so the comparison isolates the predicted geometry.
Predicted depth is defined up to an unknown scale, so each backbone is rescaled once such that its median depth matches the camera rig radius.
We evaluate on \emph{walk} with $33$ frames and truncate refinement to $24$k steps.
Truncation reduces the cost of the sweep, but the ranking is not stable at shorter budgets, so we report the spread across backbones rather than their order.
PSNR is foreground-masked and reported on the three evaluation splits of \mpNine: Interp.$_A$ on held cameras at held timestamps, Recon.\ on held cameras at supervised timestamps, and Interp.$_B$ on supervised cameras at held timestamps.

\begin{table}[tb]
\centering
\caption{\textbf{Feed-forward backbone ablation} on \emph{walk}, refinement truncated to $24$k steps. Each backbone supplies only depth. Foreground-masked PSNR (dB) on the splits of \mpNine.}
\label{tab::backbone}
\setlength{\tabcolsep}{3pt}
\begin{tabular}{@{}l ccc@{}}
\toprule
Backbone & Interp.$_A\uparrow$ & Recon.$\uparrow$ & Interp.$_B\uparrow$\\
\midrule
VGGT~\cite{vggt}                       & 20.82 & 24.99 & 20.62\\
Depth Anything 3~\cite{da3} & 20.79 & 25.24 & 20.89\\
$\pi^3$~\cite{pi3}                     & 21.09 & 25.47 & 20.97\\
MapAnything~\cite{mapanything}         & 21.02 & 25.32 & 20.86\\
\midrule
\textbf{\textcolor{linkblue}{VGGT-$\Omega$~\cite{vggtomega}, ours}} & \textbf{\textcolor{linkblue}{20.91}} & \textbf{\textcolor{linkblue}{25.45}} & \textbf{\textcolor{linkblue}{20.89}}\\
\bottomrule
\end{tabular}
\end{table}

\noindent\textbf{Results.}
Table~\ref{tab::backbone} reports quality after refinement on the three evaluation splits.
The five backbones span $0.30$~dB, which falls within the $\pm0.3$~dB run-to-run variation of our protocol, so no backbone is statistically separated from the others.
The spread remains small despite substantial differences in model capacity and output resolution: the backbone with the lowest depth resolution, $168{\times}224$ compared with up to $392{\times}518$, is not the weakest of the five.
Refinement therefore depends on the structural correctness of the anchor rather than on its metric accuracy.
An approximately correct geometry that avoids the coordinate degeneracy of unanchored initialization is sufficient, and all five backbones satisfy this requirement.
This is consistent with the from-scratch collapse reported in \mpExp, which is caused by the absence of a geometric reference rather than by an inaccurate one.
We adopt VGGT-$\Omega$ for the main experiments because it supports CPU offloading to $\peakoff$~GB, as required by the deployment budget, and not because it attains the highest score.

\noindent\textbf{Exclusion of feed-forward splat predictors.}
A second class of feed-forward networks predicts Gaussians directly rather than depth~\cite{charatan2024pixelsplat,chen2024mvsplat,tang2024lgm}, with recent methods targeting sparse-view novel-view synthesis.
These methods assume a static scene observed by a moving camera and estimate geometry from inter-view parallax.
Our anchor input has the opposite configuration: a stationary camera observing a moving subject, where inter-view differences encode motion rather than parallax.
Evaluating these models on such input violates their geometric assumption, and the resulting scores would reflect this mismatch rather than model quality.
A fair comparison requires a different anchor protocol, with one multi-view reconstruction per timestamp, which we leave outside the scope of this ablation.

\section{The Transmitted Package from Free Viewpoints}
\label{sec::apppkg}
\begin{figure}[!htb]
\centering
\includegraphics[width=\columnwidth]{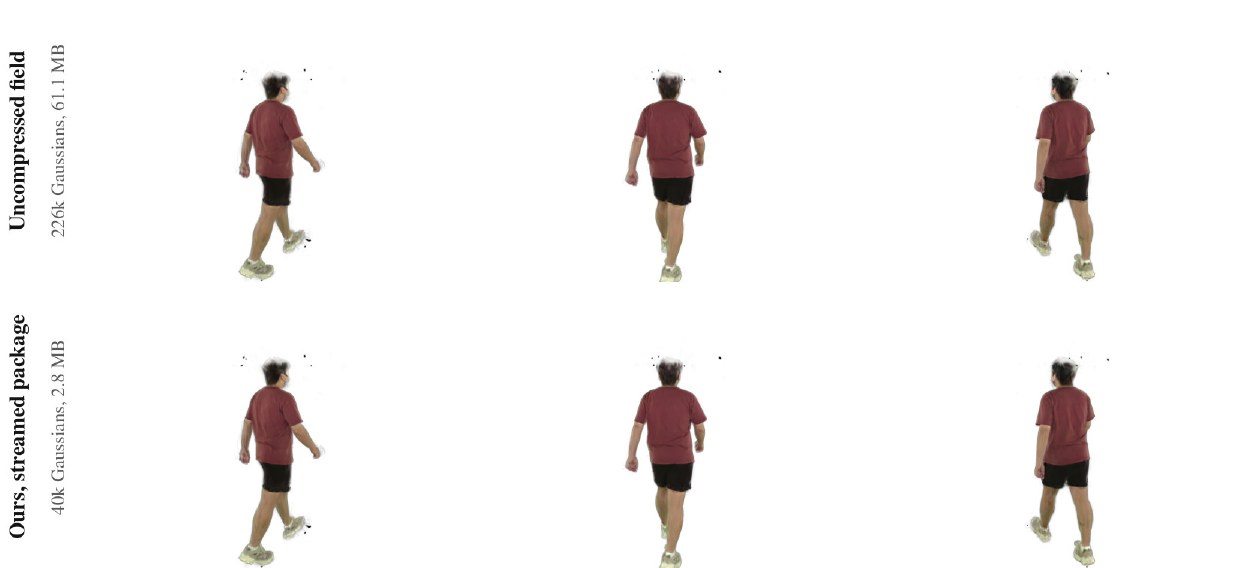}
\caption{\textbf{The transmitted package from free viewpoints.} \emph{walk}, before and after pruning and coding.}
\label{fig::webxr}
\end{figure}

Held-out evaluation is defined at the capture cameras, so it does not report what a headset shows when the viewer walks around the subject. Figure~\ref{fig::webxr} renders the transmitted package from three viewpoints outside the capture rig. At the deployment budget the package is $22\times$ smaller than the raw field and is not visually distinguishable from it; the WebGL client decodes it in shader at $\questfps$~fps on a Meta Quest Pro.

\begin{figure}[!htb]
\centering
\includegraphics[width=\columnwidth]{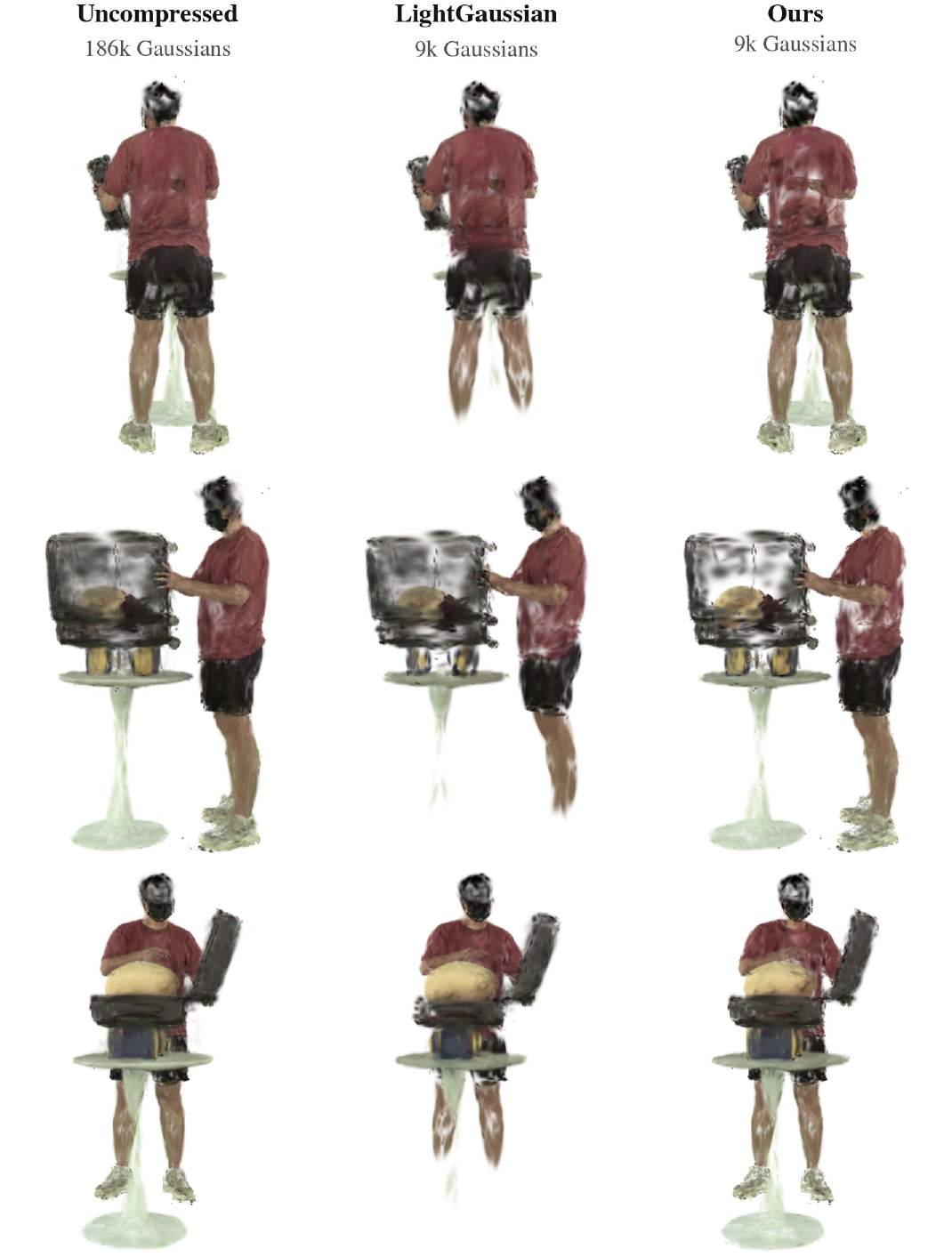}
\caption{\textbf{Client decoding at the deployment budget.} \emph{opencase}, three viewpoints outside the capture rig, all decoded by the same WebGL client that runs on the headset. Columns share the viewpoint, the timestamp and the coder; only the ranking that selected the retained Gaussians differs.}
\label{fig::ondevice}
\end{figure}

Figure~\ref{fig::ondevice} repeats this comparison against a pruning baseline on \emph{opencase}, with the retained set fixed to the deployment budget for both.
The three columns are produced by loading three package files into one client build, so the renderer, the decoder and the camera are identical and the only difference is which Gaussians survived.
At this budget the baseline drops the feet and the lower half of the carried case, which are thin structures supported by few Gaussians, and the client renders the gaps as translucent stumps.
Our package retains them.
The failure is not a loss of texture detail but a loss of parts, which is what the topology floor is designed to prevent and what a viewer notices immediately when walking around the subject.

\section{Degradation Along the Budget Ladder}
\label{sec::appladder}
\begin{figure*}[!tb]
\centering
\includegraphics[width=\textwidth]{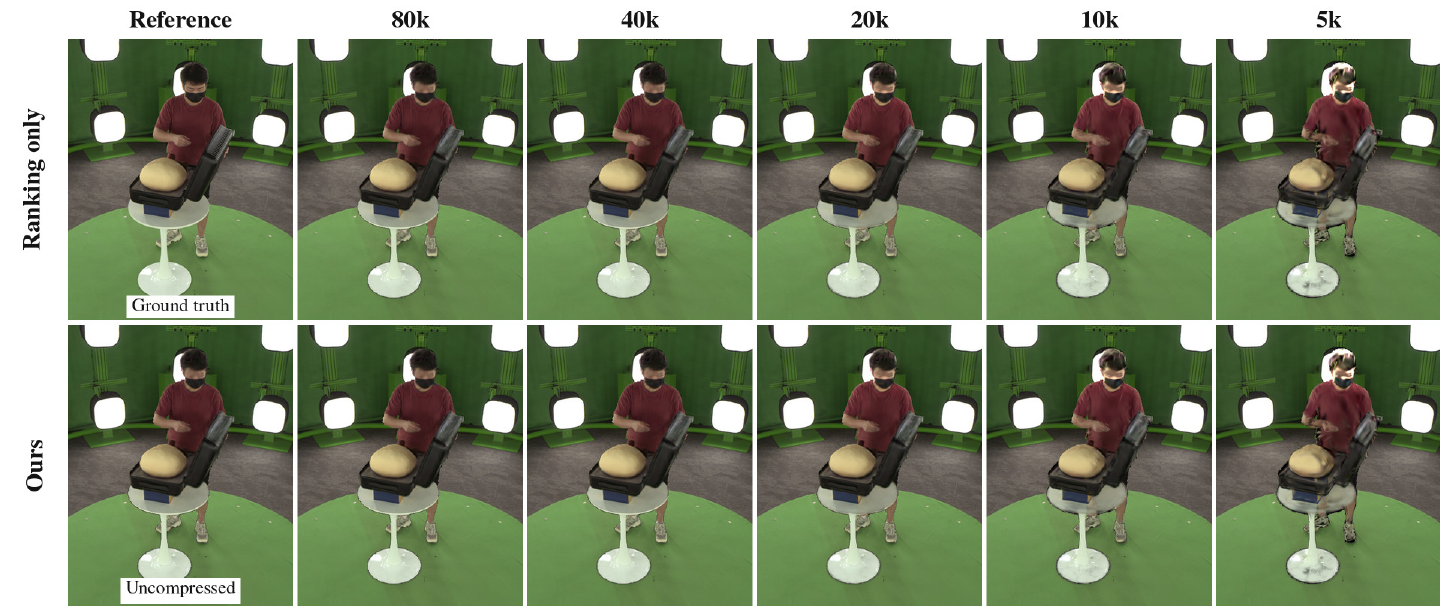}
\caption{\textbf{A package along the budget ladder.} \emph{opencase} at a held camera and a held timestamp. Columns are Gaussian budgets; the leading column is the reference. Both rows use the same ranking, the same coder and the same starting field.}
\label{fig::ladder}
\end{figure*}

\mpTfs reports the quality of a package as a number per budget.
Figure~\ref{fig::ladder} shows what those numbers look like.
The top row prunes by the importance ranking alone and the bottom row adds the verified topology floor, with everything else held fixed.
The two rows are indistinguishable at the loose budgets, where the ranking alone already retains every part of the subject.
They separate as the budget tightens: the ranking-only row loses the face and the top of the carried case first, because those regions are covered by many small Gaussians of individually low importance, and the loss is abrupt rather than gradual.
The floor spends part of the same budget on the Gaussians that carry those components and degrades their appearance instead of removing them.

\section{Reconstruction on All Nine Scenes}
\label{sec::appnine}
\begin{figure*}[!tb]
\centering
\includegraphics[width=\textwidth]{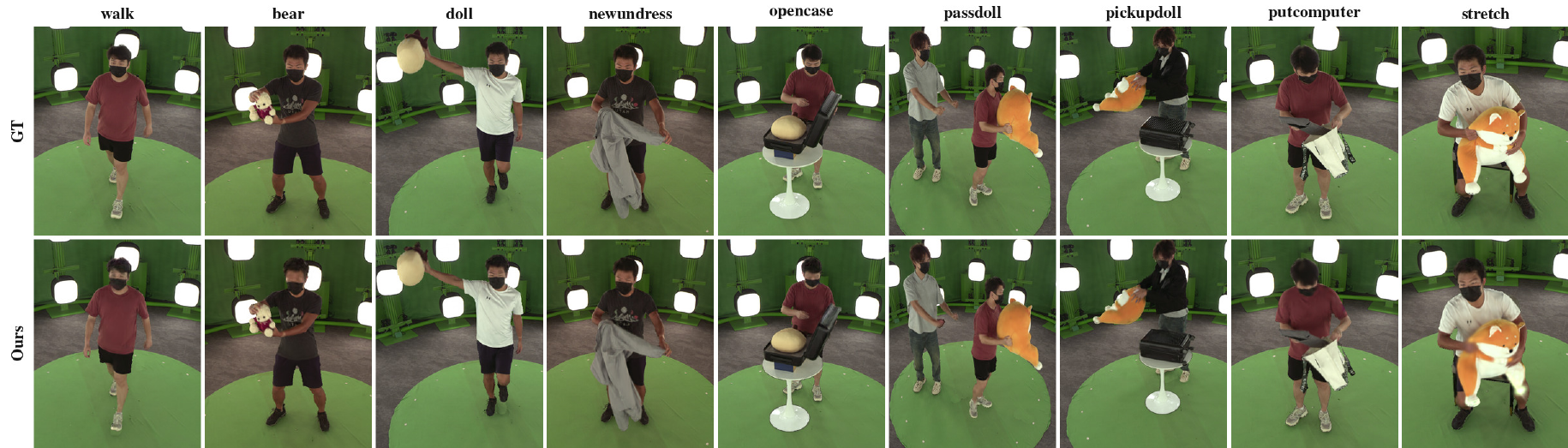}
\caption{\textbf{All nine scenes at a held camera and a held timestamp.} Crops are centred on the subject; the field is composited over the person-free plate of the same camera, as in \mpQual.}
\label{fig::ninequal}
\end{figure*}

Figure~\ref{fig::ninequal} shows the refined field on every scene of \mpNine at a held camera and a held timestamp, so no pixel in the bottom row was supervised.
The scenes cover locomotion, two-person interaction, object manipulation and loose clothing.
Reconstruction is faithful for the body and for rigid carried objects across all nine.
The residual errors are concentrated where they are expected: the trailing edges of loose fabric, where the cubic trajectory smooths motion that is not polynomial, and the contact regions between two subjects, where the anchor supplies a single depth per ray and the mirrored back side of one subject falls inside the other.

\section{Comparison with Baselines on Further Scenes}
\label{sec::appbaselines}
\begin{figure*}[!tb]
\centering
\includegraphics[width=0.78\textwidth]{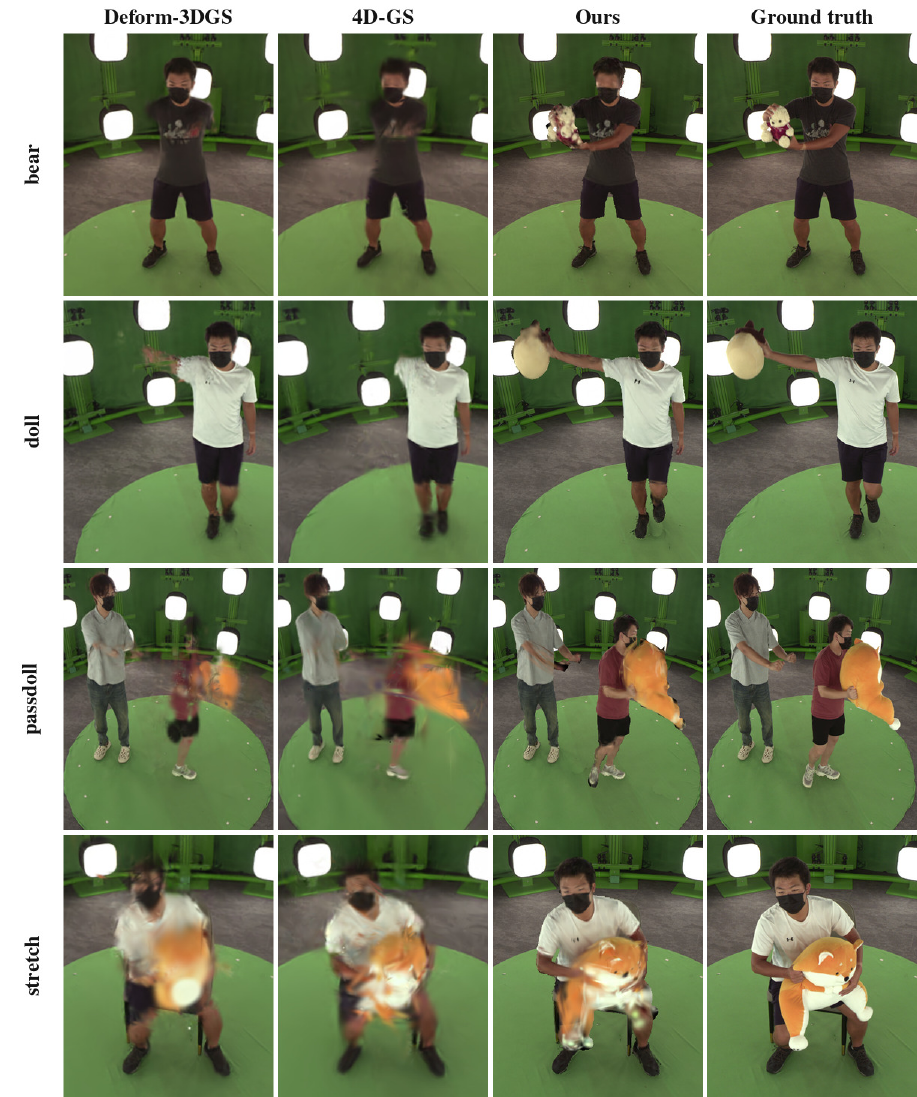}
\caption{\textbf{Held-timestamp interpolation on four further scenes.} One held camera; no method was supervised on this timestamp. Crops are centred on the subject.}
\label{fig::morebaselines}
\end{figure*}

\mpQual compares methods on a single scene.
Figure~\ref{fig::morebaselines} repeats the comparison on four more, at a held timestamp on a held camera, using the renders the two published implementations write for their own test split.
The failure is the same in every scene and is temporal rather than spatial.
Both baselines reconstruct the static parts of the subject, and both lose whatever moves fastest: the toy in \emph{bear} disappears, the thrown object in \emph{doll} smears into the background, and the two interacting subjects of \emph{passdoll} blend into a single translucent volume.
A deformation field queried at an unobserved timestamp returns an average of the poses it was trained on, and averaging is what these renders show.
Our trajectories are evaluated at the timestamp itself, so the moving parts stay separate.

\section{Rendering Between Captured Frames}
\label{sec::apptime}
\begin{figure*}[!tb]
\centering
\includegraphics[width=\textwidth]{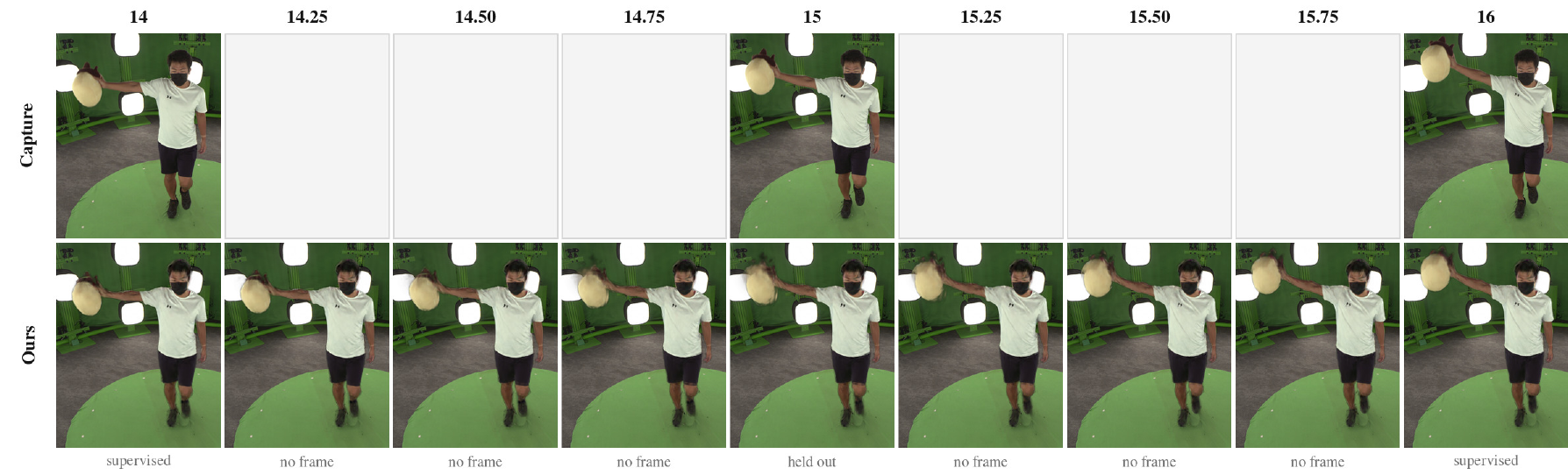}
\caption{\textbf{Quarter-frame rendering across one interval} of \emph{doll}. Top: the capture, blank where no frame exists. Bottom: our render at every timestamp.}
\label{fig::subframe}
\end{figure*}

The representation is a function of continuous time, so it can be evaluated at timestamps that the capture never recorded.
Figure~\ref{fig::subframe} renders one interval at quarter-frame steps.
Only the two endpoints were supervised; the midpoint is the held-out frame used for evaluation, and the six remaining timestamps have no corresponding capture at all.
The carried object advances by a consistent increment at every step and no pose is repeated, so the motion between captured frames is interpolated rather than snapped to the nearest supervised state.
This is the property that the held-timestamp protocol measures at one point per interval and that a viewer sees continuously during slow-motion playback.

\section{Residual Error Modes}
\label{sec::appfail}
\begin{figure*}[!tb]
\centering
\includegraphics[width=0.72\textwidth]{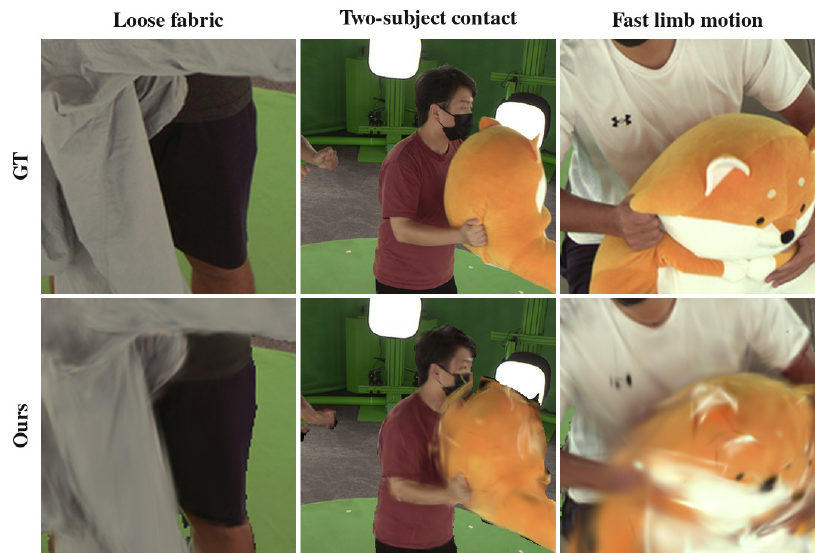}
\caption{\textbf{Residual error modes} at a held timestamp on a held camera. Crops from \emph{newundress}, \emph{passdoll} and \emph{stretch}.}
\label{fig::failure}
\end{figure*}

Figure~\ref{fig::failure} shows the three error modes that remain, as zoom crops against ground truth at a held timestamp.
Loose fabric loses its trailing edge, because a cubic trajectory cannot follow a fold that changes direction within one interval.
Contact between two subjects is reconstructed as a translucent mixture, because the anchor supplies a single depth per ray and the mirrored back side of the nearer subject is placed inside the farther one.
An object carried through a fast arc is blurred along its path, which is the same limitation as the first mode at a larger amplitude.
All three are properties of the motion model rather than of the budget, and none of them is repaired by keeping more Gaussians.

\section{Cross-Dataset Results on DNA-Rendering}
\label{sec::appdna}
Every experiment in the body uses one capture system.
This section checks that the pipeline runs on another one: DNA-Rendering~\cite{dnarendering}, an independent multi-view human capture with a different rig geometry, a different camera count, portrait rather than landscape framing, and subjects in garments the Stage-Capture scenes do not contain.

\noindent\textbf{Protocol.}
We take three sequences from the released Part~1 and convert them into the same directory layout as the body experiments.
The capture stores camera-to-world extrinsics together with lens distortion, so we undistort the images once and record the resulting pinhole intrinsics; the stored 3D keypoints reproject inside the person masks under this reading, which fixes the convention.
From the $48$-camera ring we take $32$ evenly spaced cameras and the first $33$ frames, giving the same $32{\times}33$ grid and the same every-other-frame hold-out as \mpNine.
The recipe is the one of \mpExp: the same anchor, capacity floor, loss weights and step count, and the same rule for the densification window, which is the long one where the Gaussian count settles and the short one where it does not.
The single adjustment is the shape of the anchor ray grid, which follows the portrait aspect ratio of this capture.

\noindent\textbf{Lens distortion.}
Part of this capture is calibrated with a high-order radial model whose polynomial stops increasing inside the frame.
Undistorting such a view with a standard rectification map folds the region beyond that radius back onto itself and duplicates image content into a border ring, which enters the person masks as foreground the reconstruction cannot explain.
We therefore blank each view outside the radius where its polynomial remains monotonic.
Ten of the $32$ views of one sequence require this and none of the other two do; leaving the folded ring in place inflates the foreground of an affected view by half and makes the optimization diverge.
The baselines are the same two public implementations, trained on the same converted sequences, and all numbers come from the evaluator used for the body tables.

\begin{figure*}[!p]
\centering
\includegraphics[width=0.66\textwidth]{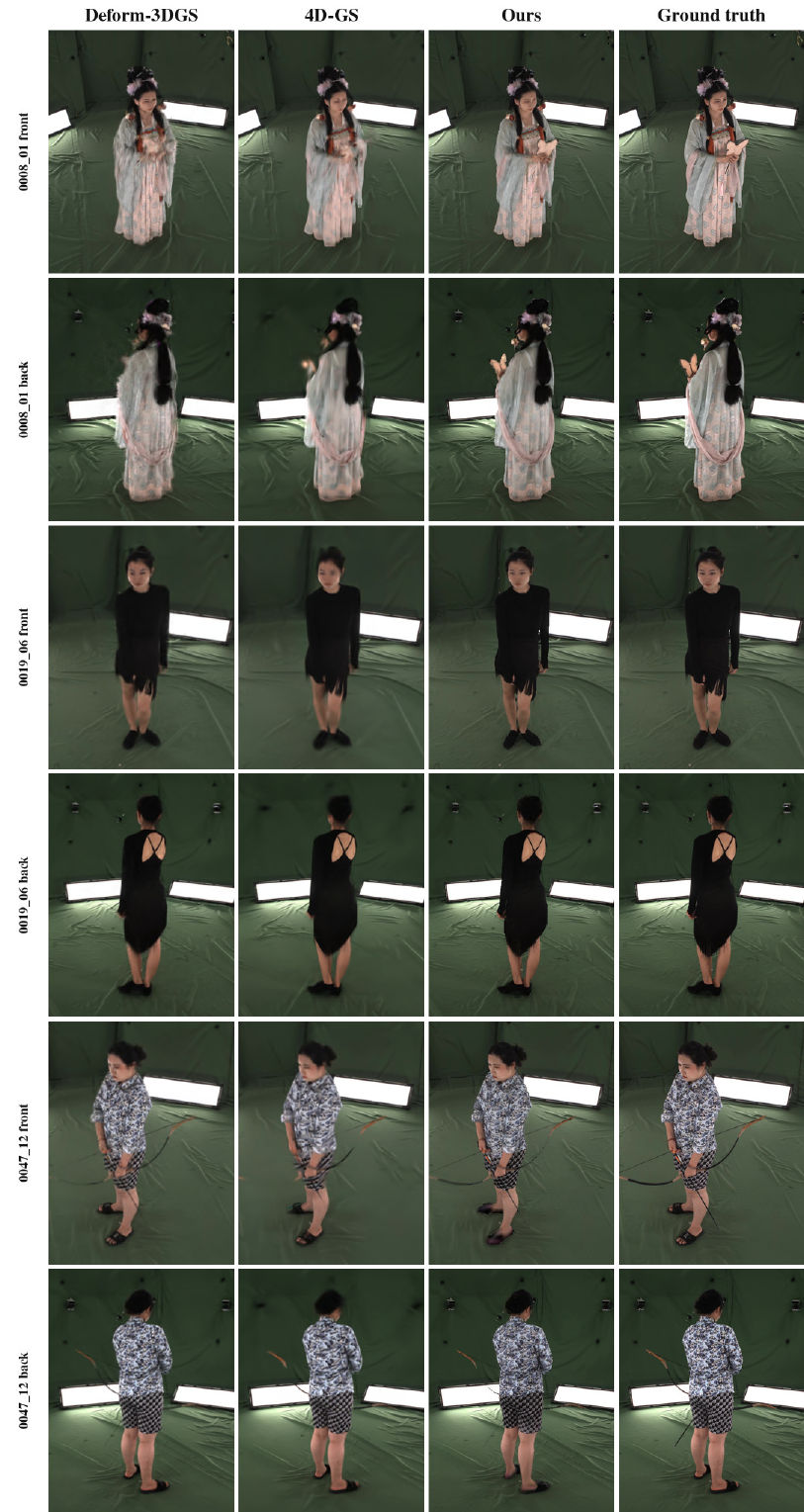}
\caption{\textbf{Held-timestamp interpolation on DNA-Rendering.} Two held cameras per sequence, front and back; no method was supervised on this timestamp. Crops are centred on the subject.}
\label{fig::dnaqual}
\end{figure*}

\noindent\textbf{Results.}
Table~\ref{tab::dna} reports held-timestamp PSNR per sequence and Figure~\ref{fig::dnaqual} shows the same sequences at a held timestamp.
Our mean is $\dnamean$~dB against $\dnadgsmean$~dB for Deformable-3DGS and $\dnafourmean$~dB for 4D-GS, from fields averaging $\dnags{\times}10^{4}$ Gaussians fitted in $\dnatime$~minutes.
We are ahead on two of the three sequences and behind 4D-GS by about a decibel on \emph{0019\_06}, the sequence on which every method scores highest.
Absolute values are not comparable with \mpNine, since the subject covers a different fraction of the frame and the masks come from a different segmentation system.
The claim here is only that the pipeline runs on a capture it was never tuned on and reconstructs it at a quality comparable to methods trained on the same sequences, with the anchor supplying usable geometry for a rig and a wardrobe it has not seen.

\begin{table*}[!tb]
\centering
\caption{\textbf{Cross-dataset results on DNA-Rendering.} Foreground-masked held-timestamp PSNR (dB) per sequence, with the mean SSIM and LPIPS of the same split. Protocol and evaluator as in \mpNine.}
\label{tab::dna}
\setlength{\tabcolsep}{4pt}
\begin{tabular}{@{}l ccc c cc@{}}
\toprule
Method & 0008\_01 & 0019\_06 & 0047\_12 & Mean$\uparrow$ & SSIM$\uparrow$ & LPIPS$\downarrow$\\
\midrule
Deformable-3DGS~\cite{yang2024deformable} & 20.02 & 27.60 & 22.24 & 23.29 & 0.984 & 0.017\\
4D-GS~\cite{wu20244d} & 22.70 & 29.56 & 21.09 & 24.45 & 0.984 & 0.021\\
\textbf{\textcolor{linkblue}{Ours}} & \textbf{\textcolor{linkblue}{24.31}} & \textbf{\textcolor{linkblue}{28.57}} & \textbf{\textcolor{linkblue}{22.72}} & \textbf{\textcolor{linkblue}{25.20}} & \textbf{\textcolor{linkblue}{0.987}} & \textbf{\textcolor{linkblue}{0.013}}
\\
\bottomrule
\end{tabular}
\end{table*}

\end{document}